\documentclass[final,5p,times,twocolumn]{elsarticle}

\usepackage{amssymb}

\newcommand{\bm}[1]{\mbox{\boldmath ${#1}$}}

\usepackage{multirow}
\usepackage{color}
\usepackage{colortbl}

\usepackage{lineno}

\journal{Computer Vision and Image Understanding}

\begin{document}

\begin{frontmatter}



\title{Semi- and Weakly-supervised Human Pose Estimation}


\author[label1]{Norimichi Ukita\corref{cor1}}
\author[label2]{Yusuke Uematsu}

\address[label1]{Toyota Technological Institute, Japan}
\address[label2]{Nara Institute of Science and Technology, Japan}

\cortext[cor1]{Corresponding author. address: 2-12-1, Tempaku, Nagoya
  468-8511, Japan, phone: +81-52-809-1832, email: ukita@ieee.org. He
  worked at Nara Institute of Science and Technology formerly.}

\begin{abstract}
  For human pose estimation in still images, this paper proposes three
  semi- and weakly-supervised learning schemes.  While recent advances
  of convolutional neural networks improve human pose estimation using
  supervised training data, our focus is to explore the semi- and
  weakly-supervised schemes.  Our proposed schemes initially learn
  conventional model(s) for pose estimation from a small amount of
  standard training images with human pose annotations.  For the first
  semi-supervised learning scheme, this conventional pose model
  detects candidate poses in training images with no human
  annotation. From these candidate poses, only true-positives are
  selected by a classifier using a pose feature representing the
  configuration of all body parts.  The accuracies of these candidate
  pose estimation and true-positive pose selection are improved by
  action labels provided to these images in our second and third
  learning schemes, which are semi- and weakly-supervised learning.
  While the first and second learning schemes select only poses that
  are similar to those in the supervised training data, the third
  scheme selects more true-positive poses that are significantly
  different from any supervised poses. This pose selection is achieved
  by pose clustering using outlier pose detection with Dirichlet
  process mixtures and the Bayes factor.  The proposed schemes are
  validated with large-scale human pose datasets.
\end{abstract}


\end{frontmatter}


\section{Introduction}
\label{section:introduction}

Human pose estimation is useful in various applications including
context-based image retrieval, etc.
The number of given training data has a huge impact on
pose estimation
as well as various recognition problems (e.g, general object
recognition \cite{DBLP:conf/nips/KrizhevskySH12} and face recognition
\cite{DBLP:conf/cvpr/TaigmanYRW14}),
Although the scale of datasets for human pose estimation has been
increasing (e.g., 305 images in the Image Parse dataset in 2006
\cite{DBLP:conf/nips/Ramanan06}, 2K images in the LSP dataset
\cite{DBLP:conf/bmvc/JohnsonE10} in 2010, and around 40K human poses
observed in 25K images in the MPII human pose dataset
\cite{DBLP:conf/cvpr/AndrilukaPGS14} in 2014), it is difficult to
develop a huge dataset for human pose estimation in contrast to object
recognition (e.g., over 1,430K images in ISVRC2012--2014
\cite{ILSVRCarxiv14}).
This is because human pose annotation
is much complicated
than weak label and window annotations for object recognition.

\begin{figure}[t]
  \begin{center}
    \includegraphics[width=\columnwidth]{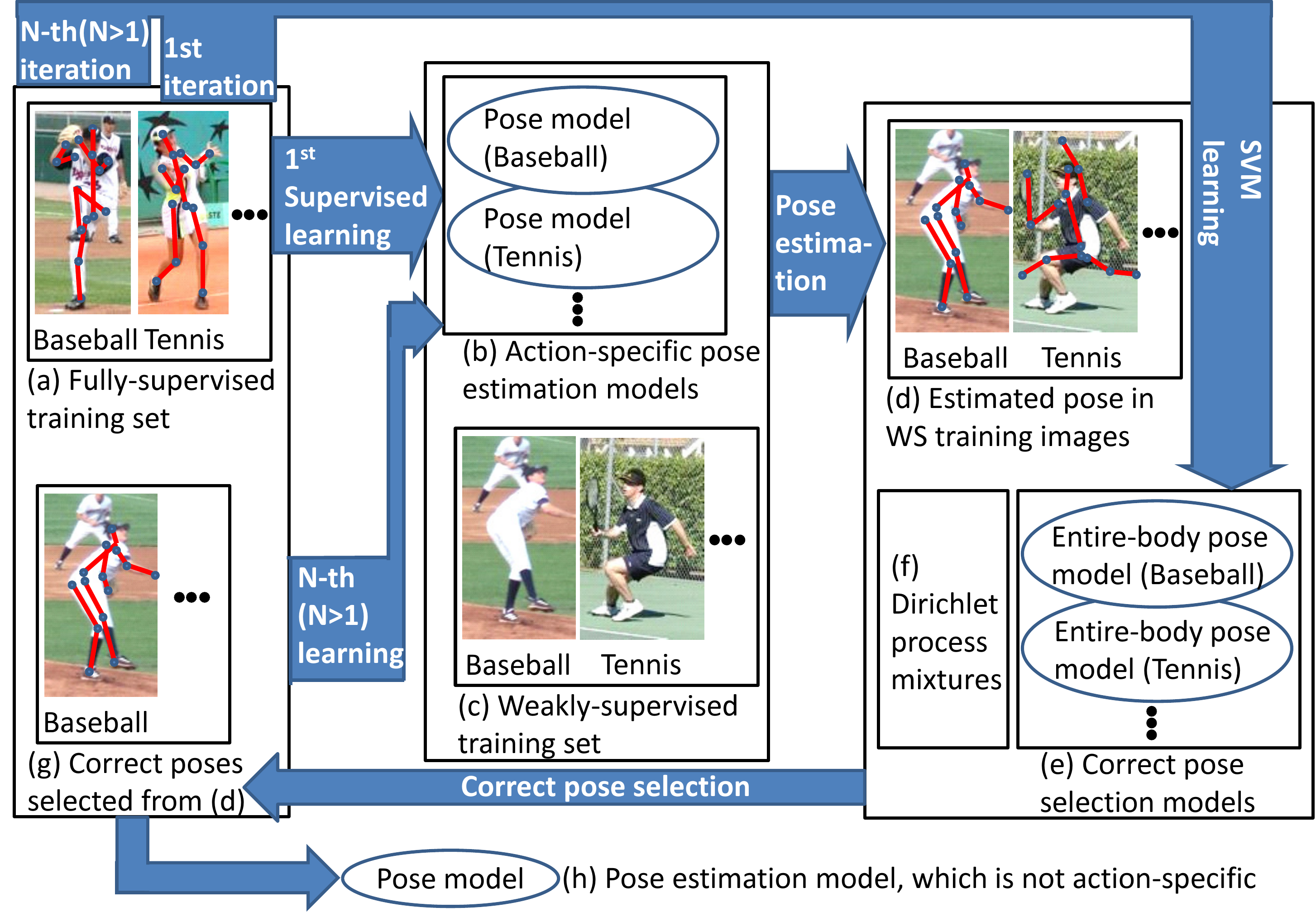}
    \caption{Overview of the proposed method.
      Images each of which has a pose annotation and an action label
      (i.e., Figure \ref{fig:top} (a)) are used
      to acquire initial action-specific pose models (i.e.,
      (b)). Each model is acquired from images along with its
      respective action label.
      $a$-th action-specific pose model is used for estimating
      human poses in training images with the label of $a$-th action
      (i.e., (c)).
      Each estimated pose (i.e., (d)) is evaluated
      whether or not it is true-positive. This evaluation is achieved by a
      pose feature representing the configuration of all body
      parts (i.e., (e)).
      True-positive poses are selected also by outlier detection using
      Dirichlet process mixtures (i.e., (f)).
      These true-positive poses are
      employed as pose annotations (i.e., (g)) for
      re-learning the action-specific pose
      models. After iterative re-learning,
      all training images with pose annotations (i.e.,
      (a) and (g)) are used for learning a final
      pose model (i.e., (h)).
      While this paper proposes three learning schemes,
      described in Sections \ref{section:semi}, \ref{section:weakly}, and
      \ref{section:weak_clustering}, this figure illustrates the third
      one, which contains all functions proposed in
      Sections \ref{section:semi}, \ref{section:weakly}, and
      \ref{section:weak_clustering}.}
    \label{fig:top}
  \end{center}
\end{figure}

To increase the number of training images
with less annotation cost,
semi- and weakly-supervised learning schemes are applicable.
Semi-supervised learning allows us to automatically provide
annotations for a large amount of data based on a small amount of
annotated data. In weakly-supervised learning, only simple annotations
are required in training data and are utilized to acquire full
annotations in learning.

We apply semi- and weakly-supervised learning to human pose estimation,
as illustrated in Figure \ref{fig:top}. In our
method with all functions proposed in this paper,
fully-annotated images each of which has a pose annotation (i.e.,
skeleton) and an action label are used to acquire initial pose models
for each action (e.g., ``Baseball'' and ``Tennis'' in
Figure \ref{fig:top}).
These action-specific pose models are used to estimate candidate
human
poses in each action-annotated image. If a
candidate pose is considered true-positive, the given pose with its
image is used for re-learning the corresponding action-specific pose
model.

The key contributions of this work are threefold:
\begin{itemize}
\item 
  True-positive poses are selected from candidate poses based on a
  pose feature representing the configuration of all body parts.
  This is in contrast to a pose estimation step in which only the
  pairwise configuration of neighboring/nearby parts is evaluated for
  efficiency.
\item The action label of each training image is utilized for
  weakly-supervised learning. Because the variation of human poses in
  each action is smaller, pose estimation in each action works better
  than that in arbitrary poses.
\item A large number of candidate poses are clustered by
  Dirichlet process mixtures for selecting true-positive poses based
  on the Bayes factor.
\end{itemize}

\section{Related Work}
\label{section:related}

A number of methods for human pose estimation employed (1) deformable
part models (e.g., pictorial structure models
\cite{DBLP:journals/ijcv/FelzenszwalbH05}) for globally-optimizing an
articulated human body and (2) discriminative learning for optimizing
the parameters of the models
\cite{DBLP:journals/pami/FelzenszwalbGMR10}.
In general, part connectivity in a deformable part model is defined by
image-independent quadratic functions for efficient optimization via
distance transform.
Image-dependent functions (e.g.,
\cite{DBLP:conf/eccv/SappTT10,DBLP:conf/cvpr/Ukita12}) disable
distance transform but improve pose estimation accuracy.
In \cite{DBLP:conf/nips/ChenY14}, on the other hand, image-dependent
but quadratic functions enable distance transform for representing the
relative positions between neighboring parts.

While global optimality of the PSM is attractive, its ability to
represent complex relations among parts and the expressive power of
hand-crafted appearance feature
are limited compared to deep
neural networks.
Recently, deep convolutional neural networks (DCNNs) improve human
pose estimation as well as other computer vision tasks.
While DCNNs are applicable to the PSM framework in order to represent
the appearance of parts as proposed in \cite{DBLP:conf/nips/ChenY14},
DCNN-based models can also model the distribution of body parts.
For example, a DCNN can directly estimate the joint locations
\cite{DBLP:conf/cvpr/ToshevS14}.
In \cite{DBLP:conf/nips/TompsonJLB14}, multi-resolution DCNNs are
trained jointly with a Markov random field.
Localization accuracy of this method
\cite{DBLP:conf/nips/TompsonJLB14} is improved by coarse and fine
networks
in \cite{DBLP:conf/cvpr/TompsonGJLB15}.
Recent approaches explore sequential structured estimation to iteratively improve the joint locations~\cite{DBLP:conf/eccv/RamakrishnaMHBS14,DBLP:conf/cvpr/SinghHF15,carreira2016human,wei2016cpm}.
%
\textcolor{black}{
One of these methods, Convolutional pose machines \cite{wei2016cpm},
is extended to
real-time pose estimation of multiple people \cite{cao2017realtime}
and hands \cite{simon2017hand}.
Ensemble modeling can be also applied to DCNNs for human pose
estimation \cite{DBLP:journals/cviu/KawanaUHY18}.
Pose estimation using DCNNs is also extended to a variety of
scenarios such as personalized pose estimation in videos
\cite{DBLP:conf/cvpr/CharlesPMHZ16} and 3D pose estimation with
multiple views \cite{pavlakos17harvesting}.
}
As well as DCNNs accepting image patches, DNNs using multi-modal
features are applicable to human pose estimation; multi-modal features
extracted from an estimated pose (e.g., relative positions between
body parts) are fed into a DNN for refining the estimated pose
\cite{DBLP:conf/cvpr/OuyangCW14}.

While aforementioned advances improve pose estimation demonstrably,
all of them require human pose annotations
(i.e., skeletons annotated on an image)
for supervised learning.
Complexity in time-consuming pose annotation work leads to
annotation errors by crowd sourcing,
as described in \cite{DBLP:conf/cvpr/JohnsonE11}.
For reducing the time-consuming annotations in supervised learning,
semi- and weakly-supervised learning are widely used.

Semi-supervised learning allows us to utilize a huge number of
non-annotated images
for various recognition problems (e.g., human action recognition
\cite{DBLP:conf/cvpr/JonesS14a}, human
re-identification\cite{DBLP:conf/cvpr/LiuSTZCB14},
and face and gait recognition \cite{DBLP:journals/tcsv/HuangXN12}).
In general, semi-supervised learning annotates the images
automatically by employing several cues in/with the images; for
example, temporal consistency in tracking
\cite{DBLP:conf/iccv/LiQHPJ11}, clustering
\cite{DBLP:conf/cvpr/MahmoodMO14}, multimodal keywords
\cite{DBLP:conf/cvpr/GuillauminVS10}, and domain adaptation
\cite{DBLP:conf/cvpr/JainL11}.

For human pose estimation also, several semi-supervised learning
methods have been proposed. However, these methods are designed for
limited simpler problems.
For example, in
\cite{DBLP:conf/bmvc/NavaratnamFC06,DBLP:conf/cvpr/KanaujiaSM07}, 3D
pose models representing a limited variation of human pose sequences
(e.g., only walking sequences) are trained by semi-supervised learning;
in \cite{DBLP:conf/bmvc/NavaratnamFC06} and
\cite{DBLP:conf/cvpr/KanaujiaSM07}, GMM-based clustering and manifold
regularization are employed for learning unlabeled data, respectively.
For
semi-supervised learning, not only a small number of annotated images
but also a huge amount of synthetic images (e.g., CG images with
automatic pose annotations) are also useful with transductive learning
\cite{DBLP:conf/iccv/TangYK13}.

In weakly-supervised learning, only part of
full annotations are given manually.
In particular, annotations that can be easily annotated are given.
For human activities, full annotations may include
the pose, region, and attributes (e.g., ID, action class) of
each person. Since it is easy to provide the attributes rather than
the pose and region, such attributes are often given
as weak annotations.
For example, only an action label is given to each training sequence
where the regions of a person
(i.e. windows enclosing a human body) in frames
are found automatically in
\cite{DBLP:conf/eccv/ShapovalovaVCLM12}.
Instead of the manually-given action label, scripts are employed as
weak annotations in order to find correct action labels of several
clips in video sequences in \cite{DBLP:conf/iccv/DuchenneLSBP09};
action clips are temporally localized.
Not only in videos but also in still images, weak annotations can
provide highly-contextual information. In
\cite{DBLP:journals/pami/PrestSF12}, given an action label, a human
window is spatially localized with an object used for this action.
For human pose estimation, Boolean geometric relationships between
body parts are used as weak labels in
\cite{DBLP:conf/cvpr/Pons-MollFR14}.

Whereas pose estimation using only action labels is more difficult
than human window localization described above, it has been
demonstrated that the action-specific property of a human pose is
useful for pose estimation (e.g. latent modeling of dynamics
\cite{DBLP:conf/iccv/UkitaHK09,DBLP:journals/cviu/UkitaK12}, switching
dynamical models in videos \cite{DBLP:conf/cvpr/ChenKWJ09},
efficient particle distribution in multiple pose models in videos
\cite{DBLP:conf/eccv/GallYG10,DBLP:journals/ivc/Ukita13},
and pose model selection in still images
\cite{bib:ukita:action_pose}).

\section{Human Pose Estimation Model}
\label{section:pose}

This section introduces two base models for human pose estimation,
deformable part models and DCNN-based heatmap models.

\subsection{Deformable Part Models}
\label{subsection:psm}

\textcolor{black}{
A deformable part model is an efficient model for articulated pose
estimation
\cite{DBLP:journals/ijcv/FelzenszwalbH05,DBLP:journals/pami/FelzenszwalbGMR10,DBLP:conf/nips/ChenY14}.
A tree-based model is defined by a set of nodes, $\bm{V}$, and a
set of links, $\bm{E}$, each of which connects two nodes.
Each node corresponds to a body part and has pose parameters (e.g.,
2D image coordinates,
orientation, and scale), which localize the respective parts.
Pose parameters are optimized by maximizing the given score function
consisting of a unary term, $S^{i}(\bm{p}_{i})$ , and a pairwise
term, $P^{i,j}(\bm{p}_{i},\bm{p}_{j})$,
as follows:
\begin{eqnarray}
  f_{\beta} (\bm{I}, \bm{P}) & = & \sum_{i \in \bm{V}} S^{i}(\bm{p}_{i}) + \sum_{i,j \in \bm{E}}
  P^{i,j}(\bm{p}_{i},\bm{p}_{j}) \label{eq:psm}
\end{eqnarray}
where $\bm{p}_{i}$ and $\bm{P}$ denote a set of pose parameters of the
$i$-th part and a set of $\bm{p}_{i}$ of all parts (i.e.  $\bm{P} =
\left\{ \bm{p}_{1}, \cdots, \bm{p}_{N^{(V)}} \right\}^{T}$, where
$N^{(V)}$ denotes the number of nodes), respectively.
$S^{i}(\bm{p}_{i})$ is a similarity score of the $i$-th part at
$\bm{p}_{i}$.
$P^{i,j}(\bm{p}_{i},\bm{p}_{j})$ is a spring-based function with a
greater value if the relative configuration of pairwise parts,
$\bm{p}_{i}$ and $\bm{p}_{j}$, is
probable.
}

\textcolor{black}{
In a discriminative training methodology proposed in
\cite{DBLP:journals/pami/FelzenszwalbGMR10},
the parameters of functions $S^{i}(\bm{p}_{i})$ and
$P^{i,j}(\bm{p}_{i},\bm{p}_{j})$ are trained with pose-annotated
positive and negative training images. 
}

\subsection{DCNN-based Heatmap Models}
\label{subsection:dcnn}

\begin{figure}[t]
  \begin{center}
    \includegraphics[width=\columnwidth]{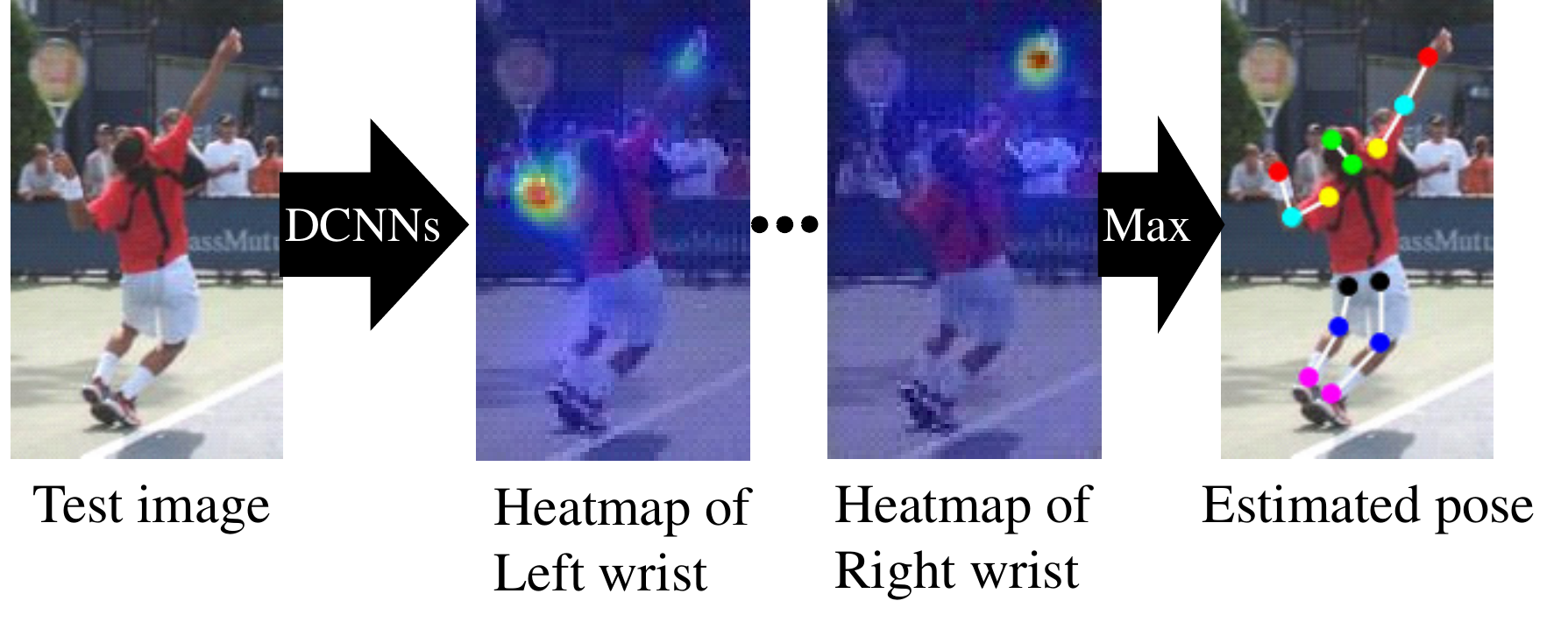}
    \caption{Common process flow of pose estimation using DCNN-based
      heatmap models.}
    \label{fig:dcnn}
  \end{center}
\end{figure}

\begin{figure*}[t]
  \begin{center}
    \includegraphics[width=\textwidth]{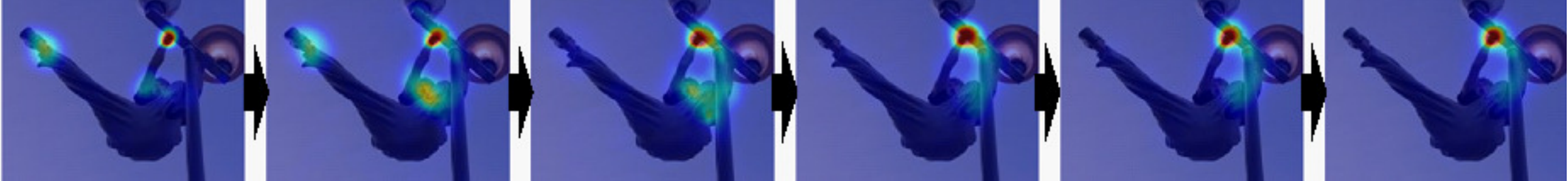}
    
    ~ \hspace*{1mm}
    1st stage \hspace*{21mm} 2nd \hspace*{21mm} 3rd \hspace*{21mm} 4th
    \hspace*{21mm} 5th \hspace*{21mm} 6th  \hspace*{21mm} ~

    \caption{Heatmaps generated in iterative inference stages in a
      DCNN-based heat map model \cite{wei2016cpm}.
      The iterative process resolves confusions
      between similar body regions
      due to local image features and obtains a strong peak in the
      latter stages.}
    \label{fig:cpm}
  \end{center}
\end{figure*}

Unlike deformable part models, recent DCNN-based human pose estimation
methods (e.g.,
\cite{newell2016eccv,yang2016end,DBLP:conf/cvpr/ChuOLW16,bulat2016pose,DBLP:conf/eccv/LifshitzFU16,wei2016cpm,pfister2015flowing})
acquire the position of each body joint from its corresponding
heatmap.
The heatmap of each joint is outputted from a DCNN as shown in Figure
\ref{fig:dcnn}.
The position with the maximum likelihood in each heatmap is considered
to be the joint position.

General DCNNs for human pose estimation consist of convolution,
activation, and pooling layers.
In order to capture local and spatially-contextual (e.g.,
kinematically-plausible) evidences for joint localization, smaller and
wider convolutional filters are used, respectively. Further contexts
are represented by sequential/iterative feedbacks of DCNN responses;
see \cite{wei2016cpm,pfister2015flowing,carreira2016human} for
example.
Figure \ref{fig:cpm} shows an example of heatmaps generated by
iterative inference stages in \cite{wei2016cpm}, which was employed as
a base model in our experiments.

\section{Semi-supervised
 Pose
  Model
  Learning by
  Correct-pose selection using  Full-body
 Pose
  Features}
\label{section:semi}

Our semi-supervised pose model learning uses two training sets.
The first
set consists of images each of which has its human pose annotation and
action label. Each image in the second set has no annotation.
The first and second sets are called the fully-supervised (FS) and
unsupervised (US) sets, respectively.

An initial pose model is learned from the FS set. This pose model is
then used for the pose estimation of images in the US set.
Note that a sole model is used in this section unlike the
action-specific models of the complete scheme illustrated in
Figure \ref{fig:top}.
All estimated poses must be classified into true-positives and
false-positives to use only true-positives for re-learning the pose
model.

\begin{figure}[t]
  \begin{center}
    \includegraphics[height=0.12\textheight]{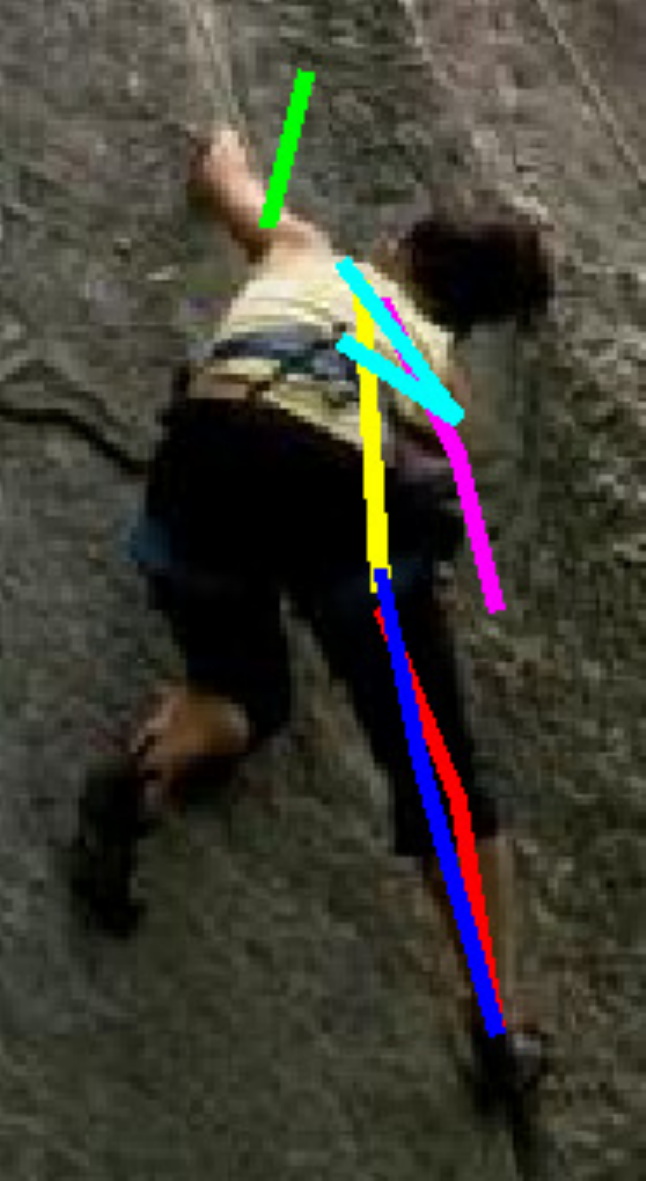}
    \includegraphics[height=0.12\textheight]{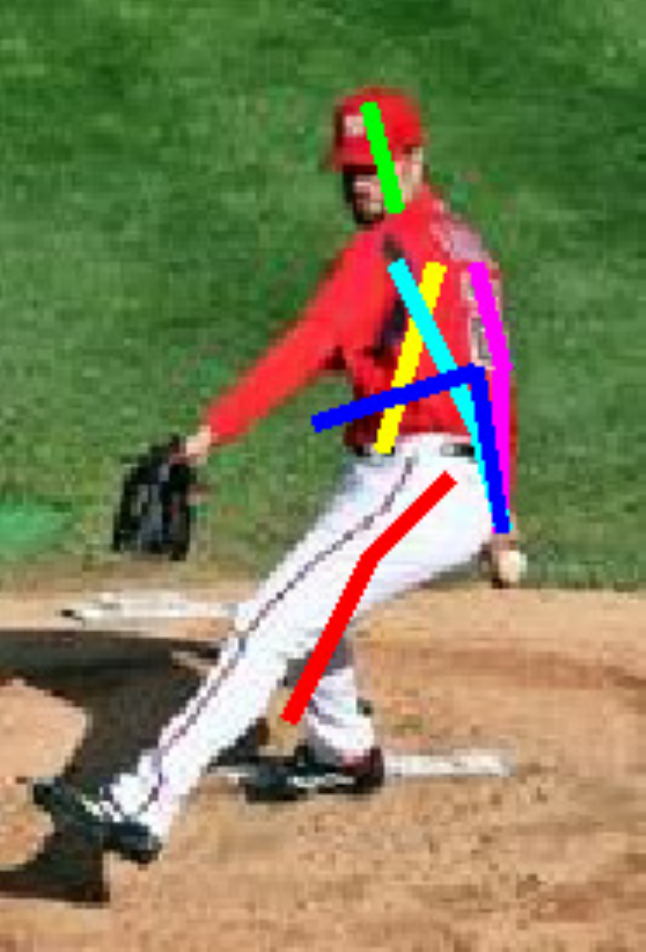}
    \includegraphics[height=0.12\textheight]{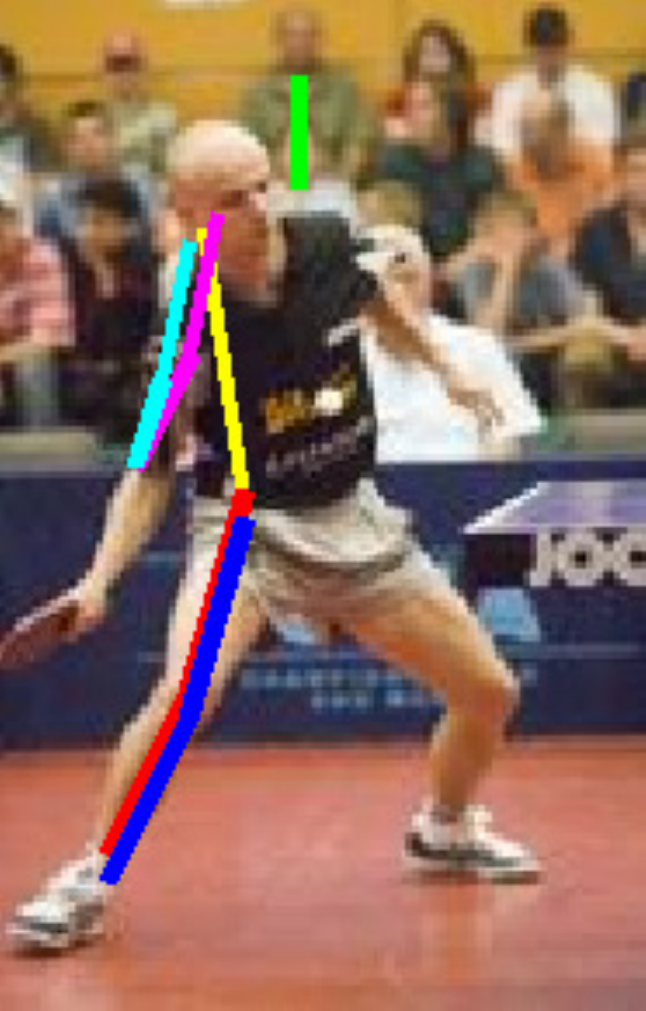}
    \includegraphics[height=0.12\textheight]{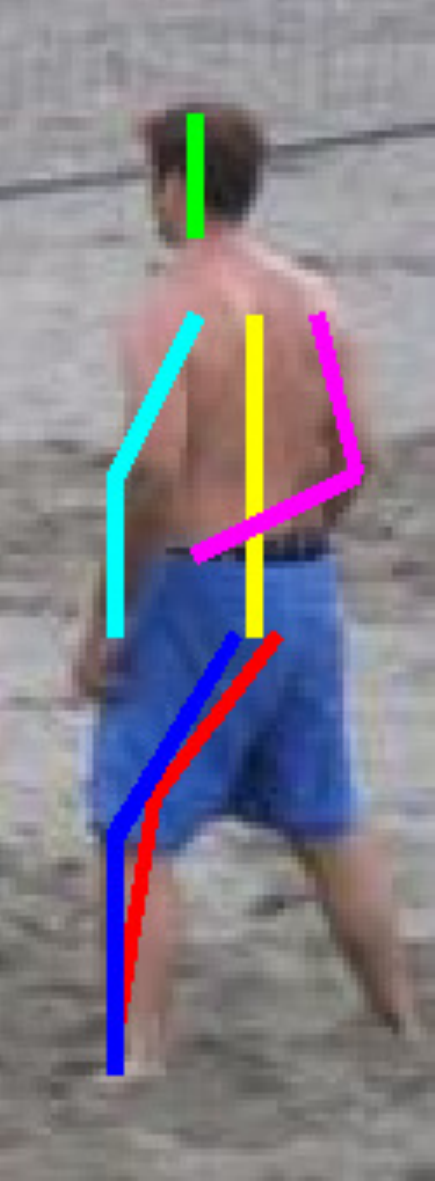}
    \includegraphics[height=0.12\textheight]{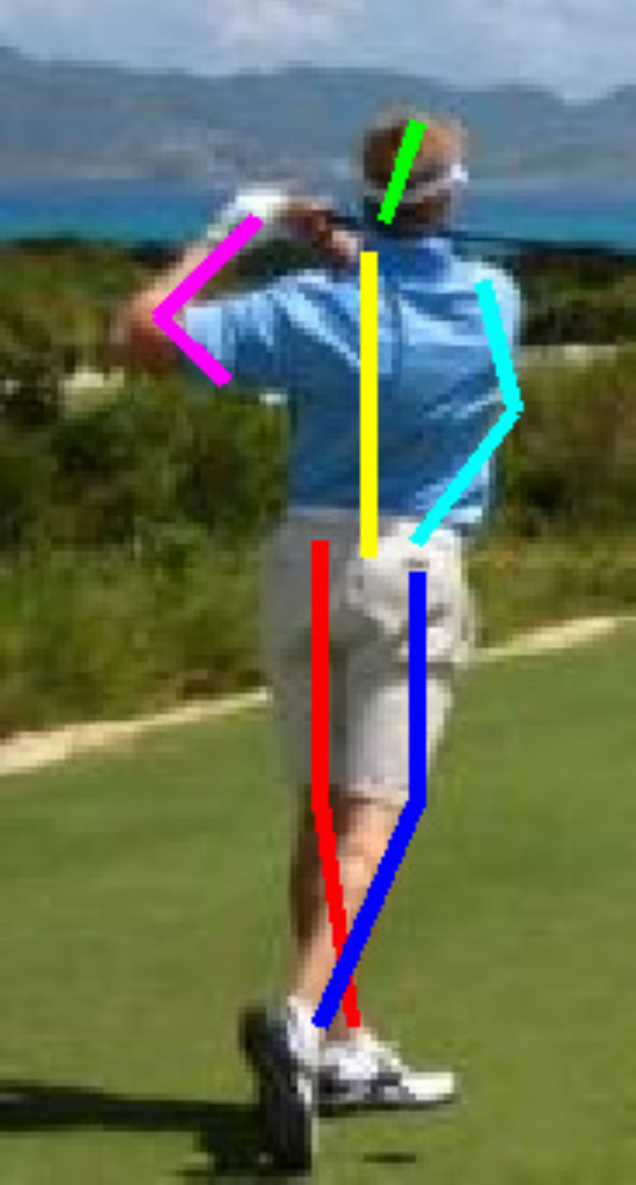}
    
    (a) \hspace*{12mm} (b) \hspace*{12mm} (c) \hspace*{12mm} (d) \hspace*{12mm} (e)

    \caption{False-positive poses estimated by the tree-based
      model \cite{DBLP:conf/cvpr/YangR11}. Green, yellow, pink,
      skyblue, red, and blue indicate the
      head, torso, right arm, left arm, right leg, and left leg,
      respectively.
      While only a few parts are incorrectly localized in (d) and (e),
      the full body is far from plausible human poses in (a), (b),
      and (c).
    }
    \label{fig:incorrect_examples}
  \end{center}
\end{figure}

Examples of false-positives are shown in
Figure \ref{fig:incorrect_examples}. Among various poses, some (e.g.,
(a), (b), and (c))
are evidently far from plausible human poses; e.g.,
the left and right limbs overlap unnaturally in (c).
Such atypical poses are obtained by a human pose model described in
Section \ref{section:pose}, because this model optimizes more or less
local regions in a full body.
Despite the relative locations of local parts being plausible, the
configuration of all parts might be implausible.

On the other hand, it is computationally possible to evaluate how
plausible each optimized configuration of the parts is after the pose
estimation process.
In our semi-supervised learning, therefore, multiple poses are
obtained from each training image in the US set by conventional pose
estimation method(s) and evaluated whether or not each of them is
plausible as the full-body configuration of a human body.
With a DPM, multiple candidate poses are obtained with a loose
threshold for score (\ref{eq:psm}).
With a DCNN-based model, all combinations of local maxima above loose
thresholds in the heatmaps are regarded as candidate poses.

\begin{figure}[t]
  \begin{center}
    \includegraphics[width=\columnwidth]{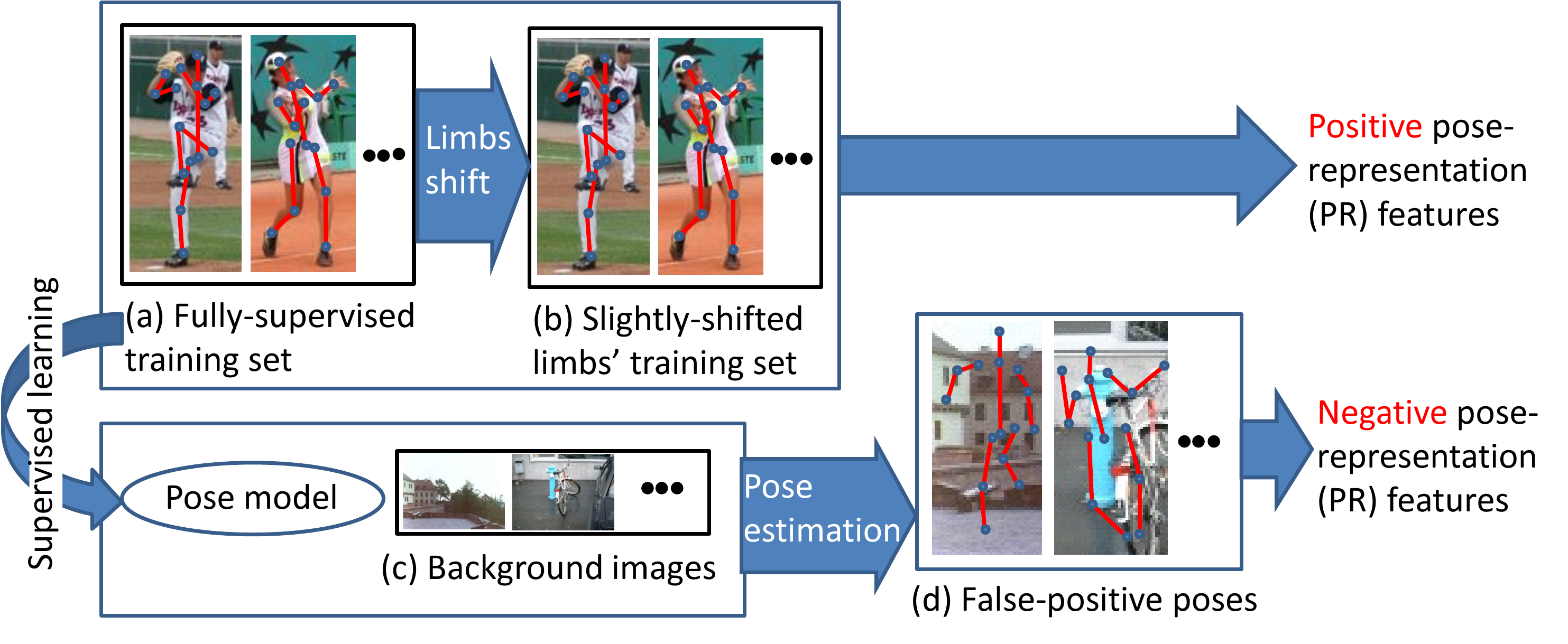}
    \caption{How to make positive and negative samples for the
      correct-pose-selection SVM (CPS-SVM).}
    \label{fig:cps_svm}
  \end{center}
\end{figure}

These candidate poses are evaluated to detect true-positive poses by
the linear SVM.
This correct-pose-selection SVM (CPS-SVM) is trained with the
following two types of samples:
\begin{description}
\item[Positive:] Images and pose annotations
  in the FS set (Figure \ref{fig:cps_svm} (a)) are used as positive samples.
  To synthesize more samples,
  in each supervised training image, the end points of all limbs in
  the pose annotation are shifted randomly (Figure \ref{fig:cps_svm}
  (b))
  \footnote{While our proposed method shifts only limbs
    to accept subtle mismatches between an estimated pose and image
    cues, a more variety of
    positive
    samples can be synthesized by image deformation according to the
    shifted limbs \cite{DBLP:conf/cvpr/PishchulinJATS12}.}
  within a predefined threshold, $\epsilon$, of the PCP evaluation
  criterion
  \cite{DBLP:conf/cvpr/FerrariMZ08,DBLP:conf/cvpr/PishchulinJATS12}.
\item[Negative:]
  Human pose estimation is applied to background images
  (Figure \ref{fig:cps_svm} (c))
  with no human region.
  Detected false-positives (Figure \ref{fig:cps_svm} (d)) are used as
  negative samples.
\end{description}

From each pose
in an image in the US set,
the following two features are extracted and concatenated to be a pose
representation (PR) feature for the CPS-SVM:
\begin{description}
\item[Configuration feature:]
\textcolor{black}{
  A PR feature should represent the
  configuration of all body parts to differentiate between different
  human poses. Such features have been proposed for action recognition
  \cite{DBLP:journals/ijcv/YaoGG12,DBLP:conf/iccv/JhuangGZSB13}. In
  the proposed method, the relational pose feature
  \cite{DBLP:journals/ijcv/YaoGG12} that is modified for 2D $x$-$y$
  image coordinates is used.
  The 2D relational pose feature \cite{DBLP:conf/iccv/JhuangGZSB13}
  consists of three components; distances between all the pairs of
  keypoints, orientations of the vector connecting two keypoints, and
  inner angles between two vectors connecting all the triples of
  keypoints. Given 14 full-body keypoints in our experiments, the
  number of these
  three components are ${}_{14}C_{2} = 91$, ${}_{14}C_{2} = 91$, and
  $3 {}_{14}C_{3} = 1092$, respectively. In total, the relational pose
  feature is a 1274-D vector.
}
\item[Appearance feature:]
  HOG features \cite{DBLP:conf/cvpr/DalalT05} are extracted from the
  windows of all parts and used
  for a PR feature.
\end{description}

The closest prior work to the CPS-SVM is presented in
\cite{DBLP:conf/eccv/JammalamadakaZEFJ12}, which is designed for
performance evaluation.
This pose evaluation method
features a marginal probability distribution for each part as well as
image and geometric features extracted from a window enclosing the
upper body.
Rather than such features, the configuration feature (i.e., 
relative positions between body parts
\cite{DBLP:journals/ijcv/YaoGG12,DBLP:conf/iccv/JhuangGZSB13}) is
discriminative between different actions and so adopted in our PR
feature; action-specific modeling is described in
Section \ref{section:weakly}.

Instead of the CPS-SVM, a DNN is used in
\cite{DBLP:conf/cvpr/OuyangCW14} for selecting true-positive poses.
Whereas DNNs are potentially powerful and actually outperform
SVM-based methods in recent
pose estimation
papers (e.g.,
\cite{DBLP:conf/cvpr/ToshevS14,DBLP:conf/eccv/RamakrishnaMHBS14,DBLP:conf/cvpr/FanZLW15,DBLP:conf/nips/TompsonJLB14,DBLP:conf/cvpr/TompsonGJLB15}),
they in general require a large number of training data for
overfit avoidance.
Since (1) our semi-supervised learning problem is assumed to have
fewer supervised training data and (2) the PR feature is a
high-dimensional data, the proposed method employs the SVM instead of
DNNs.

Detected true-positive human poses in the US set are 
then used for re-learning a pose model with the FS set.

The aforementioned pose estimation, pose evaluation, and pose model
re-learning phases can be repeated until no true-positive pose is
newly detected from the US set.
\textcolor{black}{
  In the first iteration, the pose estimation and evaluation phases
  are respectively executed with the pose model and the CPS-SVM that
  are trained by only the fully-supervised data. These pose model and
  the CPS-SVM are updated in the pose model re-learning phase and are
  used in the second or later iterations. All other settings are same
  in all iterations.
}
However, these phases were repeated only twice to avoid overfitting in
experiments shown in this paper.

This semi-supervised learning allows us to only re-learn human poses
similar to those in the FS set.
In this sense, this learning scheme is based on the smoothness
assumption for semi-supervised learning \cite{bib:Chapelle:semi}.

\section{Semi- and Weakly-supervised
 Pose
  Model
  Learning with Action-specific Pose Models}
\label{section:weakly}


\begin{figure}[t]
  \begin{center}
    \includegraphics[width=0.31\columnwidth]{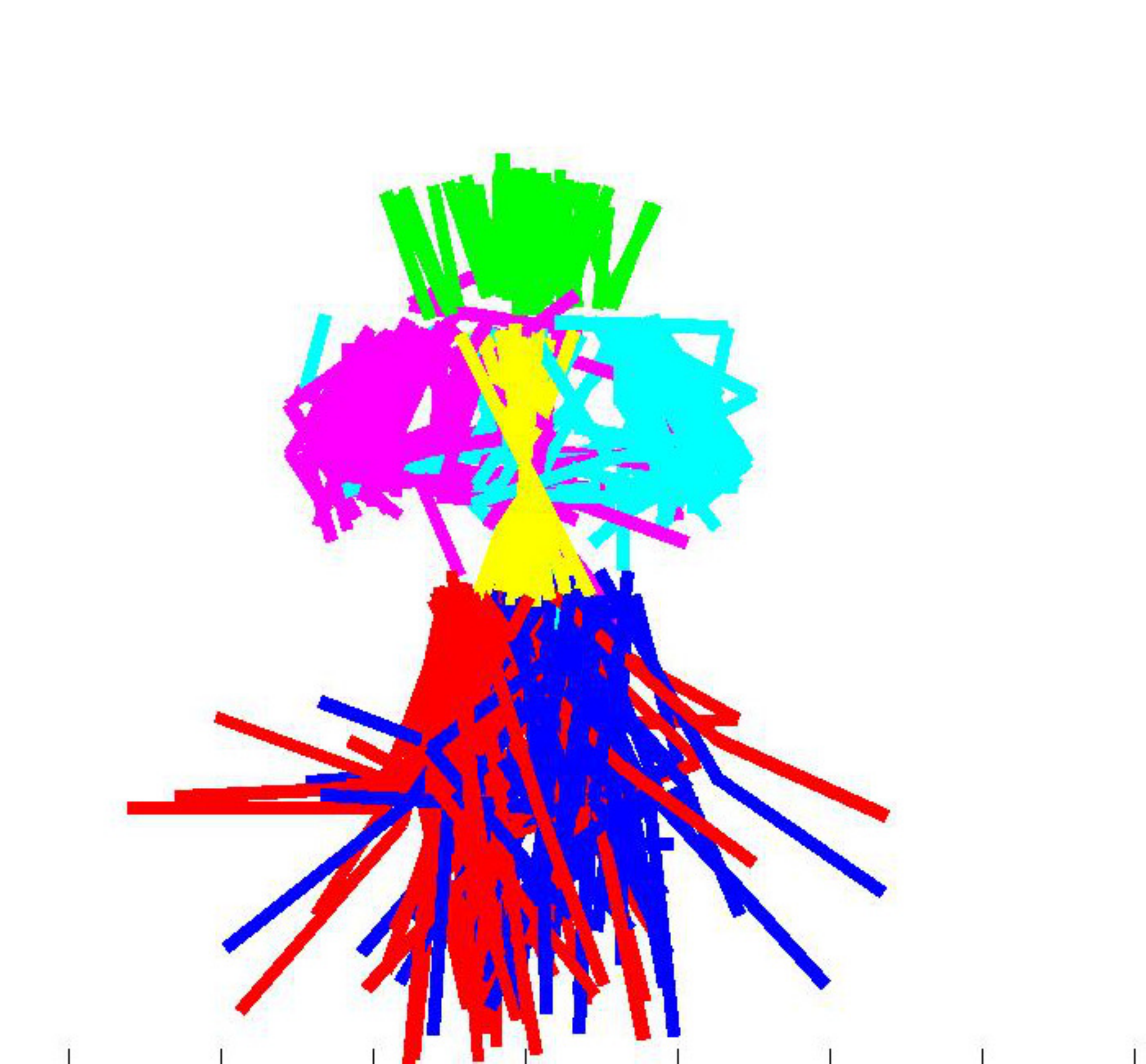}
    ~
    \includegraphics[width=0.31\columnwidth]{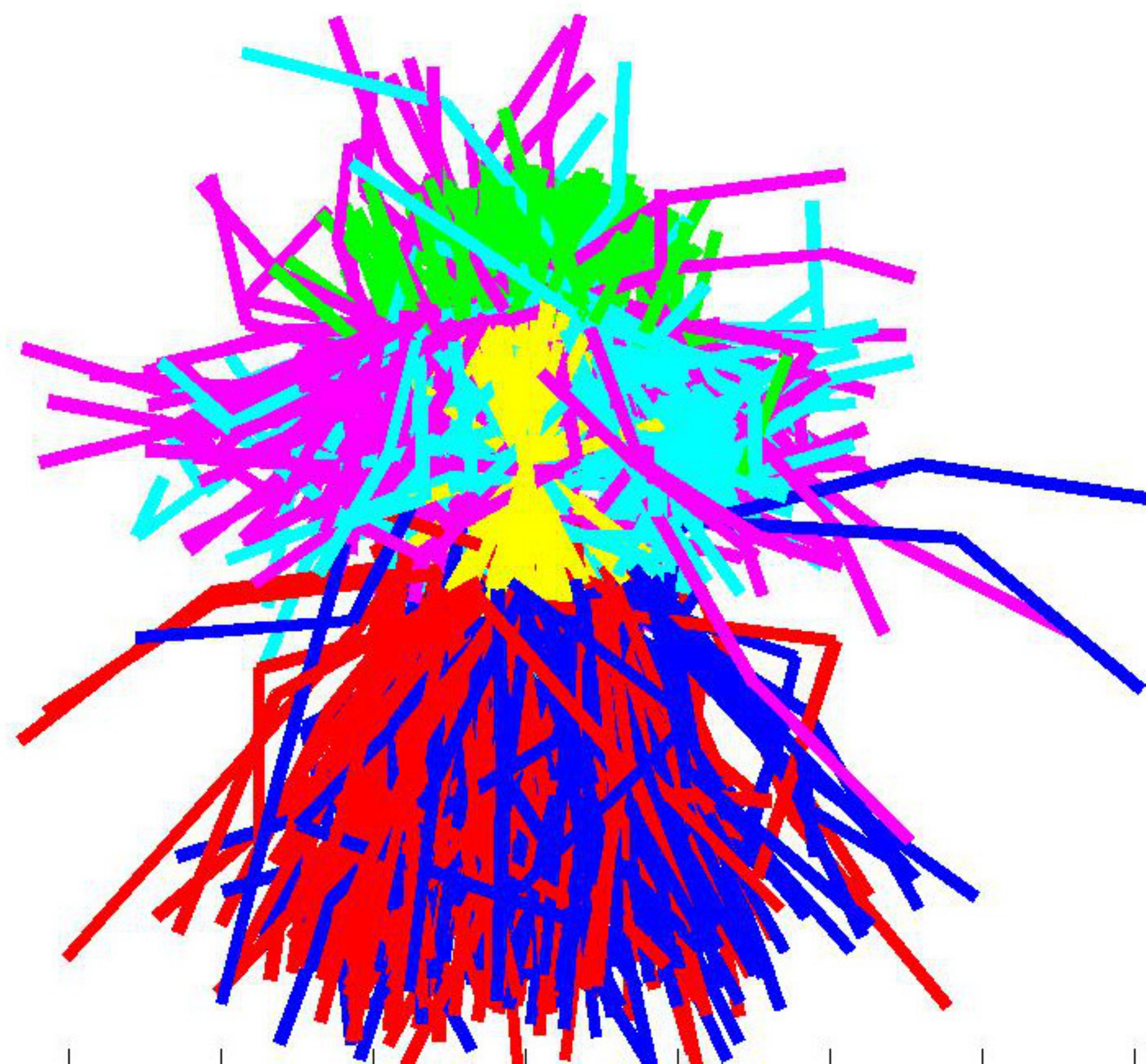}
    ~
    \includegraphics[width=0.31\columnwidth]{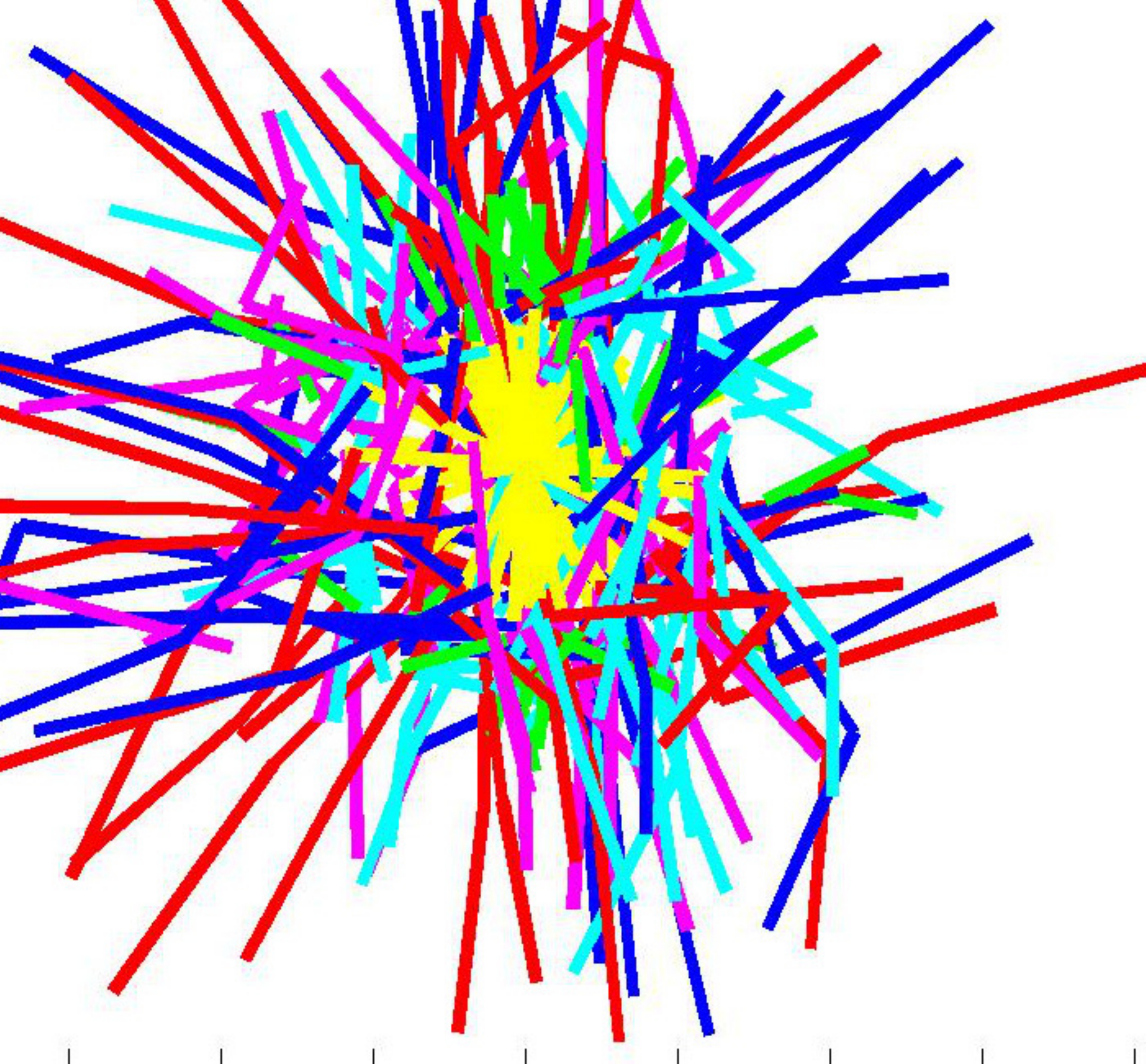}
    
    (a) athletics \hspace*{9mm} (b) badminton \hspace*{9mm} (c) gymnastics

    \caption{Pose variations based on different actions.}
    \label{fig:pose_variations}
  \end{center}
\end{figure}

In this section, semi-supervised learning, proposed in
Section \ref{section:semi}, is extended with weakly-supervised
learning.
Each image in the weakly-supervised (WS) set is
annotated with its action label.
This WS set is used for our weakly-supervised
learning instead of the US set.

The CPS-SVM proposed in the previous section is designed under the
assumption that the observed configurations of body parts are
limited. The possible configurations are more limited if the action of
a target person is known.
Figure \ref{fig:pose_variations} shows pose variations in athletics,
badminton, and gymnastics, which are included in the LSP
dataset \cite{DBLP:conf/bmvc/JohnsonE10}.
It can be seen that the pose variation depends on the action.
Based on this assumption, the proposed semi- and weakly-supervised
learning generates action-specific pose models and CPS-SVMs.
Since action-specific models are useful under the assumption of pose
clusters depending on the action, the cluster assumption
\cite{bib:Chapelle:semi} is utilized in this learning scheme.

Initially, a general pose model is learned using all training images
in the FS set. This initial model is then optimized for each
action-specific model by using only its respective images in the FS
set.
For this optimization, in a DPM pose estimation model, the general
model is used as an initial model in order to
re-optimize the parameters of two functions $S^{i}(\bm{p}_{i})$ and
$P_{i,j}(\bm{p}_{i}, \bm{p}_{j})$ in Eq. (\ref{eq:psm})
by using only the training images of each action.
In a DCNN-based model, the general model is fine-tuned using the
training images of each action.

The $a$-th action-specific pose model is used to estimate candidate
poses in images with the $a$-th action label in the WS set. Each
estimated pose is evaluated whether or not it is correct by the $a$-th
action-specific CPS-SVM.
If the estimated pose is considered correct, this pose and its
respective image are used for iterative re-learning of the $a$-th
action-specific pose model with the FS set, as with semi-supervised
learning described in Section \ref{section:semi},

After the iterative re-learning scheme finishes,
a pose model is
learned from all actions' images used in this re-learning (i.e., all images in
the FS set and WS images in which correct human poses are selected).

\section{Semi- and Weakly-supervised
 Pose
  Model
  Learning with Outlier Detection by Clustering based on Dirichlet
  Process Mixtures}
\label{section:weak_clustering}

A key disadvantage of the
learning schemes described
in Sections \ref{section:semi} and \ref{section:weakly}
is that the CPS-SVM allows us to only extract human poses similar to
those included in the FS set.
In other words, it is difficult to re-learn poses whose 2D
configurations of body parts are plausible but quite different from
those in the FS set.
The method proposed in this section extracts more true-positive
poses based on the assumption that similar true-positives
compose cluster(s) in the PR feature space of each action.

Let $N^{(P)}$ and $N^{(C)}$ denote the number of all candidate poses and
their clusters, respectively.
While $N^{(C)}$ is unknown, clustering with Dirichlet process mixtures
\cite{antoniak74}, expressed in Eq. (\ref{eq:DPM}),
or other non-parametric Bayesian clustering
can estimate $N^{(C)}$ and assign the PR features of the candidate
poses to the clusters simultaneously.
\begin{eqnarray}
  G | \{ \gamma, G_{0} \} & \sim & DP(\gamma, G_{0}), \nonumber \\
  \theta_{i} | G & \sim & G,  \nonumber \\
  x_{i} | \theta_{i} & \sim & p(x | \theta_{i}),
  \label{eq:DPM}
\end{eqnarray}
where $DP(\gamma, G_{0})$ denotes a Dirichlet process with scaling
factor $\gamma$ and base distribution $G_{0}$.

Clustering with Dirichlet process mixtures tends to produce clusters
with fewer features
\cite{DBLP:conf/nips/MillerH13,DBLP:journals/jmlr/WallachJDH10}.
These small clusters are regarded as outliers with false-positive
candidate
poses and must be removed in our re-learning scheme.
This outlier detection is achieved by \cite{ShotwellSlate2011}, which
evaluates the Bayes factor between an original set of clusters
and its reduced set generated by merging small (outlier) clusters with
other clusters.
This method \cite{ShotwellSlate2011} is superior to similar methods
because a Bayesian inference mechanism inference allows us to robustly
find small outlier clusters rather than simple thresholding (e.g.,
\cite{Escobar1995}).

This outlier detection \cite{ShotwellSlate2011} is based on modified
Dirichlet process mixtures, presented in Eq. (\ref{eq:mDPM}). In
Eq. (\ref{eq:mDPM}), parameter set $\bm{\theta} = \{ \theta_{1}, \cdots,
\theta_{N^{(P)}} \}$ in Eq. (\ref{eq:DPM}) is decomposed into two
parameters, $\bm{\phi}$ and $\bm{z}$; $\bm{\phi} = \{ \phi_{1},
\cdots, \phi_{N^{(C)}} \}$ is the set of $N^{(C)}$ unique values in
$\bm{\theta}$, and $\bm{z} = \{ z_{1}, \cdots, z_{N^{(P)}} \}$ is the
set of $N^{(P)}$ cluster membership variables such that $z_{j} = k$ if
and only if $\theta_{j} = \phi_{k}$.
Note that the number of unique values, $N^{(C)}$, is equal to the
number of pose clusters defined above.
If $\theta_{i} =
\theta_{j}$, features $x_{i}$ and $x_{j}$ are in the same cluster;
$\bm{X} = \{ x_{1}, \cdots, x_{N^{(P)}} \}$ is a set of all PR
features of the candidate poses in this paper.
\begin{eqnarray}
  p(\bm{z}) & \propto & \prod^{N^{(C)}}_{k=1} \alpha \Gamma(N^{(P)}_{k})\nonumber, \\
  \phi_{k} & \sim & G_{0}(\phi_{k}),  \nonumber \\
  x_{i} | z_{i} = k, \phi_{k}& \sim & p(x_{i} | \phi_{k}),
  \label{eq:mDPM}
\end{eqnarray}
where $p(\bm{z})$ is a prior mass function obtained by the Polya urn
scheme \cite{antoniak74}. $\Gamma(N^{(P)}_{k})$ is the gamma function
taking the number of PR features in $k$-th cluster (denoted by
$N^{(P)}_{k}$). If $z_{i} = z_{j}$, $x_{i}$ and $x_{j}$ are in the
same cluster. This model is a type of product partition models
\cite{bib:Hartigan1990}.

\textcolor{black}{
For the outlier detection, first of all,
an initial partition, $\bm{z}_{I}$, is obtained by clustering with
Dirichlet process mixtures \cite{bib:aldous:crp}.
%
Let $\mathcal{M}_{I}$ be the union of all partitions formed by
any sequence of merge operations on clusters in $\bm{z}_{I}$.
For practical use,
$\mathcal{M}_{I}$ is produced
from $\bm{z}_{I}$
by merging only small clusters having a few PR features.
}

The basic criterion of \cite{ShotwellSlate2011} for outlier detection
from $\bm{z}_{I}$
is the expense of model complexity of each partition in
$\mathcal{M}_{I}$.
Outliers can be detected by evaluating the evidence favoring a complex
model over a simpler model with no or fewer outliers.
The Bayes factor, which is used in a model selection problem, allows
us to evaluate this criterion (e.g., \cite{Bayarri2003}).
Given PR features,
the plausibilities of two models $\bm{z}_{I}$ and $\bm{z}_{m} \in
\mathcal{M}_{I}$ are evaluated by the following Bayes factor
$K_{I,m}$:
\begin{eqnarray}
  K_{I,m} & = & \frac{p(\bm{X} | \bm{z}_{I})}{p(\bm{X} | \bm{z}_{m})}
  \nonumber
\end{eqnarray}

\textcolor{black}{
A lower bound of $K_{I,m}$ supporting $\bm{z}_{I}$ rather than
$\bm{z}_{m}$ is obtained under the posterior condition and the prior
mass function $p(\bm{z}) \propto \prod^{N^{(C)}}_{k=1} \alpha
\Gamma(N^{(P)}_{k}) $ in Eq. (\ref{eq:mDPM}):
\begin{eqnarray}
  p( \bm{z}_{I} | \bm{X} ) & > & p( \bm{z}_{m} | \bm{X} ) \nonumber \\
  p( \bm{z}_{I} ) p( \bm{X} | \bm{z}_{I} ) & > &
  p( \bm{z}_{m} ) p( \bm{X} | \bm{z}_{m} ) \nonumber \\
  \frac{p(\bm{X} | \bm{z}_{I})}{p(\bm{X} | \bm{z}_{m})} & > &
  \frac{1}{\alpha^{\nu}}
  \frac{\prod^{N^{(C)}_{m}}_{k=1} \Gamma(N^{(P)}_{m,k})}{\prod^{N^{(C)}_{I}}_{k=1} \Gamma(N^{(P)}_{I,k})},
  \label{eq:condition}
\end{eqnarray}
}
where $\nu = N^{(C)}_{I} - N^{(C)}_{m}$ is the number of clusters
merged to arrive at $\bm{z}_{m}$. $N^{(P)}_{m,k}$ is the number of PR
features in $k$-th cluster of $m$-th partition.
In the proposed method, the number of PR features, $N^{(P)}_{I,k}$ and
$N^{(P)}_{m,k}$,
in inequality (\ref{eq:condition})
is weighted by score (\ref{eq:psm}) of pose detection as follows:
\begin{eqnarray}
  \frac{p(\bm{X} | \bm{z}_{I})}{p(\bm{X} | \bm{z}_{m})} & > &
  \frac{1}{\alpha^{\nu}}
  \frac{\prod^{N^{(C)}_{m}}_{k=1} \Gamma \left(
      \sum^{N^{(P)}_{m,k}}_{f=1} \sqrt{\overline{T}_{m,k,f}}
    \right)}{\prod^{N^{(C)}_{I}}_{k=1} \Gamma \left(
      \sum^{N^{(P)}_{I,k}}_{f=1} \sqrt{\overline{T}_{I,k,f}} \right)},
  \label{eq:weighted}
\end{eqnarray}
where $\overline{T}_{m,k,f}$ denotes the normalized score of $f$-th
pose in $k$-th cluster of $m$-th partition; the scores are normalized
linearly so that all of them are distributed between 0 and 1.

In inequality (\ref{eq:condition}), the left-hand side is the Bayes factor,
$K_{I,m}$, which is computed
for all possible partition pairs (i.e., $\bm{z}_{I}$ and $\bm{z}_{m}
\in \mathcal{M}_{I}$)
by the method of Basu and Chib
\cite{RePEc:bes:jnlasa:v:98:y:2003:p:224-235} in our proposed method.
The lower bound of $K_{I,m}$ is defined with parameter
$\alpha$ given in Eq. (\ref{eq:mDPM}).
To determine $\alpha$, the scale provided by Kass and
Raftery \cite{kassr95} gives us an intuitive interpretation.
%
Given $\alpha$,
only
if $\bm{z}_{I}$ satisfies inequality (\ref{eq:condition}) for all
$\bm{z}_{m} \in \mathcal{M}_{I}$, then each of merged small clusters
in $\mathcal{M}_{I}$ are detected as outliers.

In the proposed method, re-learning using the CPS-SVM is primarily
repeated for updating pose models. Then the updated pose models are
used for re-learning with
clustering and
outlier detection. This re-learning is repeated until no new training
image emerges for re-learning.
%
Note that all images in the WS set are used in the process of pose
estimation and outlier detection in all iterations.

\section{Experimental Results}
\label{section:experiments}

\begin{table*}[t]
  \begin{center}
    \caption{
      The number of human poses in each action class is shown.
      In our method, the training set of the LSP is divided into 500
      images for
      sully-supervised training and 500 images for weakly-supervised
      training, which are Train (FS) and Train (WS), respectively.
      Note that the action classes of test data in the LSP are shown
      just for information, while the action labels are not used for
      testing.
      In the MPII dataset, action classes for testing images are
      publicly unavailable.}
    \label{table:images}
    \begin{tabular}{l|l|c|c|c|c}
      \hline
      \multicolumn{2}{c|}{} & Athletics & Badminton & Baseball &Gym+Parkour \\ \hline \hline
      \multirow{3}{*}{LSP} & Train (FS)& 20 & 62 & 74 & 46 \\ \cline{2-6}
      & Train (WS) & 13 & 68 & 85 & 57 \\  \cline{2-6}
      & Test  & 46 & 127 & 137 & 128 \\ \hline
      LSP ext & Train  & 543 & 1 & 101 & 8547 \\ \hline \hline
      MPII & Train  & 284 & 83 & 208 & 205 \\ \hline \hline
      \multicolumn{2}{c|}{} & Soccer & Tennis & Volleyball & General \\ \hline \hline
      \multirow{3}{*}{LSP} & Train (FS)& 71 & 51 & 40 & 136 \\ \cline{2-6}
      & Train (WS) & 56 & 47 & 36 & 138 \\  \cline{2-6}
      & Test  & 125 &  93 &  87 & 257 \\ \hline
      LSP ext & Train  & 15 & 1 & 4 & 788\\ \hline \hline
      MPII & Train  & 149 &  176 &  105 & 26735 \\ \hline
    \end{tabular}
  \end{center}
\end{table*}

\subsection{Experimental Setting}
\label{subsection:setting}

The proposed method was evaluated with the publicly-available LSP, LSP
extended \cite{DBLP:conf/bmvc/JohnsonE10} and MPII human pose
\cite{DBLP:conf/cvpr/AndrilukaPGS14} datasets.
Images in the LSP dataset were collected
from Flickr
using eight action labels (i.e., text tags associated with each image),
namely athletics,
badminton, baseball, gymnastics, parkour, soccer, tennis, and volleyball.
However, not only one but also a number of tags including erroneous
ones are associated with each image.
On the other hand, an action label is given to each image in the MPII
dataset, but the action labels are more fine-grained (e.g.,
serve, smash, and receive in tennis) than the LSP.
These incomplete and uneven annotations make it difficult to
automatically give one semantically-valid action label to each image
in the datasets.

For our experiments, therefore, action labels shared among the
datasets were defined as follows; athletics, badminton, baseball,
gymnastics\footnote{
The MPII has no action label related to ``parkour'', and
human poses in ``parkour'' are similar to those in ``gymnastics''.
So ``parkour'' is merged to ``gymnastics'' in the LSP
in our experiments.
}, soccer, tennis, volleyball, and general.
Whilst one of these eight labels was manually associated with each of
2000 images (1000 training images and 1000 test images) in the LSP
dataset and 10000 images in the LSP extend dataset,
several fine-grained action classes in the MPII dataset were
merged to one of these eight classes\footnote{The activity IDs of the
  training data extracted from the MPII dataset are ``61, 126, 156,
  160, 241, 280, 307, 549, 640, 653, 913, 914, 983'', ``643, 806'',
  ``348, 353, 522, 585, 736'', ``328, 927'', ``334, 608, 931'', ``130,
  336, 439, 536, 538, 934'', ``30, 196, 321, 674, 936, 975'' for
  athletics, badminton, soccer, baseball, gymnastics, soccer, tennis,
  and volleyball, respectively.
}.
Table \ref{table:images} shows
the number of training and test data of the eight action classes
in each dataset.

For pose estimation, we used methods proposed in
\cite{DBLP:conf/nips/ChenY14}
and \cite{wei2016cpm}.  This is because \cite{DBLP:conf/nips/ChenY14}
and \cite{wei2016cpm} are one of state-of-the-arts using the PSM and
the DCNN, respectively, as shown in Table
\ref{table:results_lsp_pose}.
Each of the two methods \cite{DBLP:conf/nips/ChenY14,wei2016cpm}
obtained candidate poses by loose
thresholding, as described in Section \ref{section:semi}.
More specifically, in \cite{wei2016cpm}, the candidate poses were
generated from joint positions (i.e., local maxima in heatmaps) that
were extracted not only in the final output but also in all iterative
inference stages (e.g., Figure \ref{fig:cpm}).

\subsection{Quantitative Comparison}
\label{subsection:quantitative}

\begin{table*}
  \begin{center}
    \caption{
Quantitative comparison using test data in the LSP
      dataset and the strict PCK-0.2 metric \cite{DBLP:journals/pami/YangR13}.
      We used the person-centric
      annotations given in \cite{DBLP:conf/cvpr/JohnsonE11}.
      Ours-semi
      (g, l, q, and w), Ours-weak (h, m, r, and x), and Ours-weakC (i,
      n, s, and y) correspond to
      our semi-supervised learning (Section \ref{section:semi}), semi-
      and weakly-supervised learning (Section \ref{section:weakly}),
      semi- and weakly-supervised learning with outlier detection
      (Section \ref{section:weak_clustering}), respectively.
Our methods are implemented based on two different baselines, Chen \&
Yuille \cite{DBLP:conf/nips/ChenY14} (Baseline--1 in the Table) and Wei
et al. \cite{wei2016cpm} (Baseline--2 in the Table).
If the proposed method is implemented with Baseline--1/2, it is
called Ours-semi--1/2, Ours-weak--1/2, and Ours-weakC--1/2.
Each result is obtained on a different training dataset specified by
at the top of each set; {\bf LSP}, {\bf LSP+LSPext}, and {\bf
  LSP+LSPext+MPII}.
      For all of our proposed methods, the FS set consisted of only 500
      images in the LSP and all remaining images were used as the WS set.
      For reference, two baselines are evaluated also with only 500
      images in the LSP; (e) and (j).
For fair comparison in terms of the amount of the FS set,
(e) and (j) should be compared with our proposed methods. 
      In each training set, the best scores among supervised learning
      methods and
      methods that used only 500 images for the FS set are colored by
      {\bf \textcolor{red}{red}} and {\bf \textcolor{blue}{blue}}, respectively, in each
      column.}
    \label{table:results_lsp_pose}
\begin{footnotesize}
    \begin{tabular}{l|cccccccc}
      \hline
      Method & Head & Shoulder & Elbow & Wrist & Hip & Knee & Ankle & Mean \\ \hline \hline
\multicolumn{9}{l}{{\bf LSP}} \\ \hline
(a) Tompson et al. \cite{DBLP:conf/nips/TompsonJLB14}&90.6&79.2&67.9&63.4&69.5&71.0&64.2&72.3\\ \hline
(b) Fan et al. \cite{DBLP:conf/cvpr/FanZLW15}&92.4&75.2&65.3&64.0&75.7&68.3&\bf{\textcolor{red}{70.4}}&73.0 \\ \hline
(c) Carreira et al.~\cite{carreira2016human} &90.5&81.8&65.8&59.8&\bf{\textcolor{red}{81.6}}&70.6&62.0&73.1 \\ \hline
(d) Yang et al.~\cite{yang2016end} &90.6&78.1&\bf{\textcolor{red}{73.8}}&68.8&74.8&69.9&58.9&73.6 \\ \hline
(e) Baseline--1~\cite{DBLP:conf/nips/ChenY14} (HALF)&88.8&75.7&69.3&61.9&71.2&67.0&58.8&70.4\\ \hline
(f) Baseline--1~\cite{DBLP:conf/nips/ChenY14}&91.8&78.2&71.8&65.5&73.3&70.2&63.4&73.4\\ \hline
(g) Ours-semi--1&89.0&77.8&69.7&61.9&71.8&67.4&58.9&70.9 \\ \hline

(h) Ours-weak--1&89.0&78.0&70.0&61.7&72.1&67.8&58.9&71.1 \\ \hline

(i) Ours-weakC--1&91.5&78.0&\bf{\textcolor{blue}{71.1}}&62.6&73.1&68.7&\bf{\textcolor{blue}{62.3}}& 72.5\\ \hline

(j) Baseline--2~\cite{wei2016cpm} (HALF)& 89.1&75.8 &61.9&58.6 &71.7 & 64.8&58.9 &68.7 \\ \hline
(k) Baseline--2~\cite{wei2016cpm}&\bf{\textcolor{red}{93.5}} &\bf{\textcolor{red}{83.1}} &69.7 &\bf{\textcolor{red}{68.9}} &81.4 &\bf{\textcolor{red}{73.7}} &65.0 &\bf{\textcolor{red}{76.5}} \\ \hline
(l) Ours-semi--2&90.2&77.4&62.5&58.6&73.0&64.5&58.4&69.2 \\ \hline

(m) Ours-weak--2&90.7&78.0&63.2&59.2&74.1&66.3&58.4&70.0 \\ \hline

(n) Ours-weakC--2&\bf{\textcolor{blue}{92.1}}&\bf{\textcolor{blue}{82.0}}&67.6&\bf{\textcolor{blue}{63.3}}&\bf{\textcolor{blue}{78.8}}&\bf{\textcolor{blue}{70.6}}&60.5&\bf{\textcolor{blue}{73.6}} \\ \hline \hline
\multicolumn{9}{l}{{\bf LSP+LSPext}} \\ \hline
(o) Yu et al. \cite{DBLP:conf/eccv/YuZC16}&87.2&\bf{\textcolor{red}{88.2}}&\bf{\textcolor{red}{82.4}}&\bf{\textcolor{red}{76.3}}&\bf{\textcolor{red}{91.4}}&\bf{\textcolor{red}{85.8}}&78.7&\bf{\textcolor{red}{84.3}} \\ \hline
(p) Baseline--2~\cite{wei2016cpm}&\bf{\textcolor{red}{96.9}} &87.1 &80.4 &75.1 &86.5 &83.2 &\bf{\textcolor{red}{81.0}} & \bf{\textcolor{red}{84.3}}\\ \hline 

(q) Ours-semi--2&91.9&79.2&67.8&60.5&79.9&70.4&63.5&73.3 \\ \hline

(r) Ours-weak--2&93.2&81.8&69.0&61.5&83.7&71.7&64.7&75.1 \\ \hline

(s) Ours-weakC--2&\bf{\textcolor{blue}{94.0}}&\bf{\textcolor{blue}{84.4}}&\bf{\textcolor{blue}{74.7}}&\bf{\textcolor{blue}{68.7}}&\bf{\textcolor{blue}{83.0}}&\bf{\textcolor{blue}{79.8}}&\bf{\textcolor{blue}{72.1}}&\bf{\textcolor{blue}{79.5}} \\ \hline \hline
\multicolumn{9}{l}{\bf{LSP+LSPext+MPII}} \\ \hline
(t) Pishchulin et al.~\cite{pishchulin16cvpr} &97.0 &91.0 &83.8 &78.1 &91.0 &86.7 &82.0 &87.1 \\\hline
(u) Insafutdinov et al. \cite{DBLP:conf/eccv/InsafutdinovPAA16}&97.4&92.7&\bf{\textcolor{red}{87.5}}&\bf{\textcolor{red}{84.4}}&\bf{\textcolor{red}{91.5}}&89.9&87.2&90.1 \\ \hline

(v) Baseline--2~\cite{wei2016cpm}&\bf{\textcolor{red}{97.8}} &\bf{\textcolor{red}{92.5}} &87.0 &83.9 &\bf{\textcolor{red}{91.5}} &\bf{\textcolor{red}{90.8}} &\bf{\textcolor{red}{89.9}} &\bf{\textcolor{red}{90.5}} \\ \hline

(w) Ours-semi--2&92.4&80.8&70.3&65.7&82.5&73.3&68.8&76.3 \\ \hline

(x) Ours-weak--2&93.0&82.2&73.2&69.5&83.9&77.5&72.1&77.5 \\ \hline

(y) Ours-weakC--2&\bf{\textcolor{blue}{93.6}}&\bf{\textcolor{blue}{85.1}}&\bf{\textcolor{blue}{76.3}}&\bf{\textcolor{blue}{71.0}}&\bf{\textcolor{blue}{85.2}}&\bf{\textcolor{blue}{80.6}}&\bf{\textcolor{blue}{77.8}}&\bf{\textcolor{blue}{81.4}} \\ \hline
    \end{tabular}
\end{footnotesize}
  \end{center}
\end{table*}

Table \ref{table:results_lsp_pose} shows quantitative evaluation
results. Note that all methods except our proposed methods (i.e., (a)
-- (f), (j), (k), (o), (p), and (t) -- (v)) are based on supervised
learning.
The parameters of our proposed methods (i.e., $\epsilon$ of PCP
criterion for positive sample selection in the CPS-SVM
and $\alpha$ of the product partition model) were set as follows:
$\epsilon = 0.7$, which was determined empirically, and
$\alpha = \frac{1}{3}$, which was selected based on the scale of Kass
and Raftery \cite{kassr95} so that false-positives are avoided as many
as possible even with reduction of true-positives.
For all of our proposed methods (i.e., (g) -- (i), (l) -- (n), (q) --
(s), and (w) -- (y)),
the FS set consisted of 500 images in the LSP.
Remaining 500 images in the LSP were used as the WS set.
Additional WS sets
collected
from the LSP extended and the MPII were used in ``(q) -- (s)'' and
``(w) -- (y)''. While ``(q) -- (s)'' used only the LSP extended, both
of the LSP extended and the MPII were used in ``(w) -- (y)''.

Note that it is impossible to compare the proposed
methods (i.e., semi- and weakly-supervise learning) with supervised
learning methods on a completely fair basis. Even if the same set of
images is used, the amount of annotations is different between
semi/weakly-supervised and fully-supervised learning schemes.
In general, the upper limitation of expected accuracy in the proposed
method is the result of its baseline, if training data used in the
baseline is split into the FS and WS sets for the proposed method.

As shown in Table \ref{table:results_lsp_pose},
Baseline--2 \cite{DBLP:conf/nips/ChenY14} is superior to Baseline--1
\cite{wei2016cpm} in most joints. The same tendency is observed also
between our proposed methods using Baseline--1 and Baseline--2.
It can be also seen that, in all training
datasets, the complete version of the proposed method (i.e., 
Ours-weakC) is the best
among three variants of the proposed methods.
In what follows, only Ours-weakC--2 is compared with related work.

Comparison between (k) Baseline--2 \cite{wei2016cpm} and (n)
Ours-weakC--2 is fair in terms of the amount of training images, while
pose annotations in 500 images were not used in the proposed method.
The proposed method is comparable with the baseline in all joints.
Furthermore, (n) Ours-weakC--2 outperforms (j) Baseline--2 (HALF). This
is a case where the amount of annotated data is equal between the baseline
and the proposed method, while the proposed method also uses extra WS
data (i.e., remaining 500 training images in the LSP).

In experiments with two other training sets (i.e., ``LSP+LSPext'' and
``LSP+LSPext+MPII''), only the WS set increases while the FS set is
unchanged from experiments with the LSP dataset.
As expected, in these two experiments, difference in performance gets
larger between the baseline and the proposed methods than in
experiments with the LSP.
However, we can see performance improvement in the proposed methods as
the WS set increases; (n) 73.6 \% $<$ (s) 79.5 \% $<$ (y) 81.4 \% in
the mean accuracy.

\subsection{Detailed Analysis}

\paragraph{\textcolor{black}{Effectiveness in Each Action}}

\begin{table*}[t]
  \begin{center}
    \caption{\textcolor{black}{
      PCK-2.0 score of each action class on the test data of
      the LSP dataset.
      The scores of the baseline \cite{wei2016cpm}
      and our method are shown; 
      namely, (j) Baseline--2 (HALF), (v)
      Baseline--2 with LSP+LSPExt+MPII,
      and (y) Ours-weakC--2 with
      LSP+LSPext+MPII in Table \ref{table:results_lsp_pose}.
      Gain [($\alpha$) $\rightarrow$ ($\beta$)] $=
      \frac{S_{\beta}}{S_{\alpha}}
      \times 100$: $S_{\alpha}$
      and $S_{\beta}$ denote the scores of ($\alpha$) and ($\beta$),
      where $\{ \alpha, \beta \} \in \{ {\rm j}, {\rm v},
      {\rm y} \}$,
      respectively.
      Since the number of training images are inequivalent among
      action classes (see Table \ref{table:images}), Gain [(j)
      $\rightarrow$ (y)], which is
      the performance gain of our-weakC--2, is linearly-normalized by
      the number of training human poses of each action and shown
      within brackets.
    }}
    \label{table:actions}
  \begin{footnotesize}
    \begin{tabular}{l|c|c|c|c}
      \hline
      & Athletics & Badminton & Baseball &Gym+Parkour \\ \hline \hline
      (j) Baseline--2 \cite{wei2016cpm} (HALF) &68.6&66.7&67.5&69.5\\ \hline
      (v) Baseline--2 \cite{wei2016cpm} (LSP+LSPext+MPII)&91.6&90.7&90.8&91.0\\\hline
      (y) Ours-weakC--2 (LSP+LSPext+MPII)&84.8&75.4&80.2&80.5\\ \hline
      Baseline--2 Gain [(j) $\rightarrow$ (v)] \% &23.0&24.0&23.3&21.5\\ \hline
      Ours-weakC--2 Gain [(j) $\rightarrow$ (y)] \% &16.2 (0.19)&8.7 (0.57)&12.7 (0.032)&11.0 (0.0012)\\ \hline \hline
      & Soccer & Tennis & Volleyball & General \\ \hline \hline
      (j) Baseline--2 \cite{wei2016cpm} (HALF) &69.2&71.9&67.7&68.9\\ \hline
      (v) Baseline--2 \cite{wei2016cpm} (LSP+LSPext+MPII)&89.6&90.6&90.1&90.5\\ \hline
      (y) Ours-weakC--2 (LSP+LSPext+MPII) &86.4&80.2&78.3&83.9\\ \hline
      Baseline--2 Gain [(j) $\rightarrow$ (v)] \% &20.4&18.7&22.4&21.6\\ \hline
      Ours-weakC--2 Gain [(j) $\rightarrow$ (y)] \% &17.2 (0.078)&8.3 (0.037)&10.6 (0.073)&15.0 (0.0005)\\ \hline
    \end{tabular}    
  \end{footnotesize}
  \end{center}
\end{table*}

\textcolor{black}{
As discussed in Section \ref{subsection:quantitative}, the mean performance
gain in our semi- and weakly-supervised
learning is smaller than that in supervised learning; $81.4 -
68.7 = 12.7$ in Ours-weakC--2 vs. $90.5 - 68.7 = 21.8$ in Baseline--2 on average.
Here we investigate the performance gain in each action class rather
than on average.
Table \ref{table:actions} shows the PCK-2.0 score of each action
class on the test set of the LSP dataset.
We focus on the performance gains normalized by the number of training
human poses of each action class (i.e., values within brackets); the
number of human poses in each action class is shown in Table
\ref{table:images}.
A gap of the normalized performance gains between the baseline
\cite{wei2016cpm} and the proposed method (i.e., Ours-weakC--2 Gain
[(j) $\rightarrow$ (y)]) is smaller in ``gym+parkour'' and ``general''
classes than other classes.
The performance gains in other classes are better
because human poses are action-dependent and easy to be
modeled while ``gym+parkour'' and ``general'' classes
include a large variety of human poses; see Figure
\ref{fig:pose_variations} to visually confirm the pose
variations of ``athletics'' and ``gymnastics''.
}

\textcolor{black}{
Even the best action-specific gain in Ours-weakC--2 (i.e., $17.2
= 86.4 - 69.2$ in ``soccer'') is
less than the mean gain of Baseline--2 (i.e., $21.8 = 90.5 - 68.7$).
However, in contrast to the mean score of Ours-weakC--2 (i.e., $12.7 =
81.4 - 68.7$),
the best action-specific gain is closer to the mean gain
of Baseline--2.
In addition, the best action-specific gain normalized by the number of
training
data (i.e., $0.078 = 17.2/220$ in ``soccer'') is reasonable compared
with the
normalized mean gain in Baseline--2 (i.e., $0.00057 = 21.8/(500+10000+27945)$,
where the number of training human poses in the LSP, LSPext, and MPII
are 500, 10000, and 27945, respectively).
}

\textcolor{black}{
The results above validate that our proposed method works
better in actions
where a limited variety of human poses are observed.
}

\paragraph{Effectiveness of re-learning}

\begin{figure}[t]
  \begin{center}
    \includegraphics[width=\columnwidth]{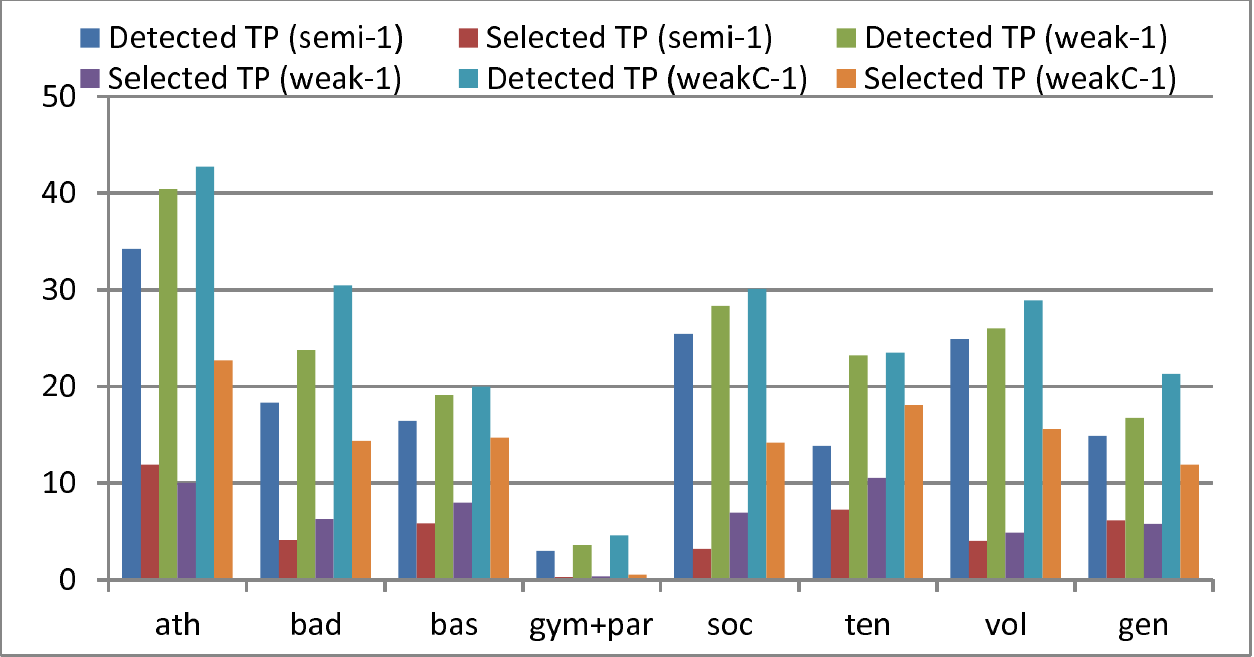}
    \caption{Quantitative evaluation of candidate pose estimation and
      true-positive pose selection in the LSP dataset. Detected TP is
      incremented if a set of candidate poses includes a true-positive
      in each image. Selected TP is incremented if a true-positive is
      selected by the CPS-SVM in each image. The vertical axis
      indicates the rate of images with detected/selected TP to all
      training images in the WS set.}
    \label{fig:pose_estimation}
  \end{center}
\end{figure}

\begin{figure}[t]
  \begin{center}
    \includegraphics[width=\columnwidth]{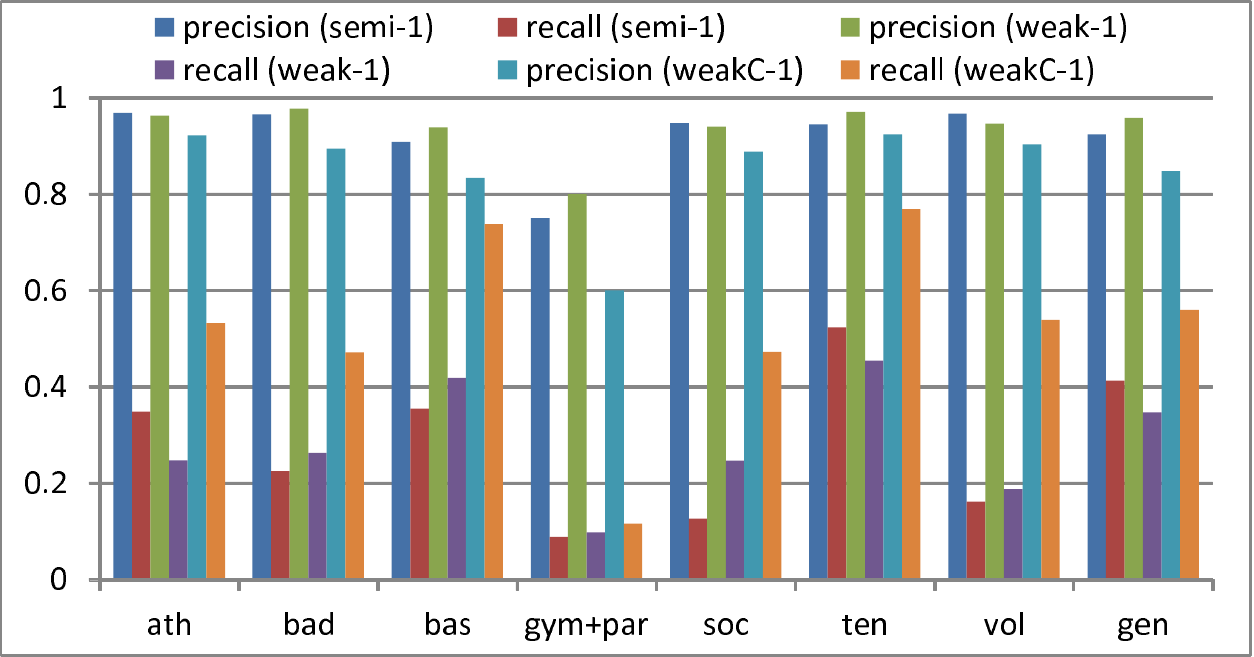}
    \caption{Quantitative evaluation of precision and recall rates for
      true-positive pose selection in the LSP dataset.}
    \label{fig:pose_selection}
  \end{center}
\end{figure}

The effectiveness of re-learning
depends on the number of true-positives selected
from the WS set. For our method using
\cite{DBLP:conf/nips/ChenY14}
with the 500 FS and 500 WS images in the LSP dataset,
Figure \ref{fig:pose_estimation} shows (1) the rate of images in which
true-positive pose(s) are included in candidate poses
(indicated by ``Detected TP'') and (2) the rate of images in which
true pose-positive(s) are correctly selected from candidate poses
(indicated by ``Selected TP'').
In this evaluation, a candidate pose is considered true-positive if
its all parts satisfy the PCP criterion
\cite{DBLP:conf/cvpr/FerrariMZ08,DBLP:conf/cvpr/PishchulinJATS12}.
Note that the results shown in Figure \ref{fig:pose_estimation} were
measured after the iterative learning ended.

In Figure \ref{fig:pose_estimation},
it can be seen that
only a few poses were selected in gymnastics even in ``Ours-weakC''.
This is natural because (1) the distribution of possible poses in
gymnastics is wide relative to the number of its training images and
(2) overlaps between two sparse distributions (i.e., poses observed in
the FS and WS sets of gymnastics) may be small.

In other actions, on the other hand, the number of selected poses
could be increased by ``Ours-weak'' and ``Ours-weakC'' in contrast to
``Ours-semi''.
The difference between the two rates (i.e., ``Detected TP'' and
``Selected TP'')
represents the number of true poses that can be selected correctly
from a set of candidate poses. This is essentially equivalent to
precision of the pose selection methods.
In addition to precision, recall is also crucial because true poses
should be selected as frequently as possible:
\begin{eqnarray}
  \rm{Precision} & = & \frac{Nmb(\bm{ATP} \cap \bm{STP})}{Nmb(\bm{STP})}, \label{eq:precision} \\
  \rm{Recall} & = & \frac{Nmb(\bm{ATP} \cap \bm{STP})}{Nmb(\bm{CP} \cap \bm{ATP})}, \label{eq:recall}
\end{eqnarray}
where $\bm{ATP}$, $\bm{STP}$, and $\bm{CP}$ denote respectively the
numbers of all true poses in the WS set (i.e., the number of images in
the WS set), candidate poses selected as true ones by pose selection
methods, and all candidate poses detected from the WS set.
$Nmb(\bm{P})$ is a function that counts the number of poses
included in pose set $\bm{P}$.
These precision and recall rates are shown in
Figure \ref{fig:pose_selection}.
From the figure, it can be seen that the precision rates are
significantly high in almost all cases. That is, most selected poses
are true-positive. Compared with the precision rates, the recall rates
are lower. That is, many false-negatives are not used for
re-learning. This means that the proposed pose selection approaches
and parameters were conservative so that only reliable poses are
selected and
used for re-learning.

\begin{figure}[t]
  \begin{center}
    \includegraphics[width=\columnwidth]{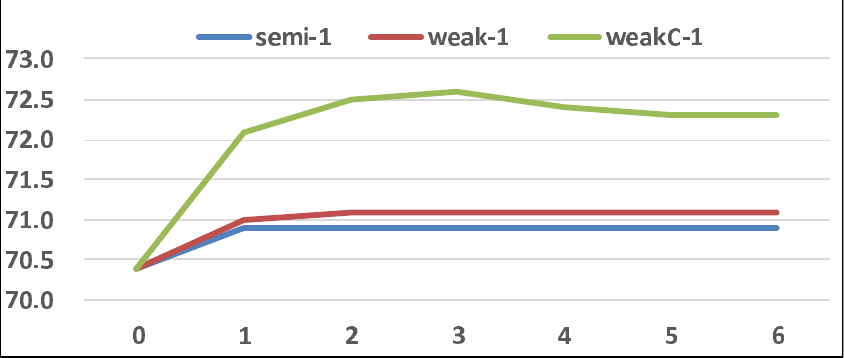}
    \caption{\textcolor{black}{
        Convergence in iterative re-learning steps on the LSP
      dataset. The vertical
      and horizontal axes indicate the PCK-0.2 score and the number of
      iterations, respectively. The 0-th iteration is executed only with the supervised training data.}}
    \label{fig:convergence}
  \end{center}
\end{figure}

\textcolor{black}{
Figure \ref{fig:convergence} shows the convergence histories of (g)
Ours-semi--1, (h) Ours-weak--1, and (i) Ours-weakC--1, which are shown
in Table \ref{table:results_lsp_pose}. 
After a big improvement in the first iteration, the second one can
also improve the score.
It can be seen that the improvement is saturated in the second iteration.
In the worst case, the score was decreased
as shown in the
third or later iterations of Ours-weakC--1.
This is caused due to overfitting and false-positive samples:
\begin{itemize}
\item The overfitting occurs when only similar samples are detected by the CPS-SVM and used for model re-learning.
The iterative sample detections using newly-detected similar samples possibly lead to detecting only similar samples.
\item If false-positives are detected by the CPS-SVM, the iterative detections using those false-positives may lead to detecting more false-positives.
\end{itemize}
}

\textcolor{black}{
Following the above results, 
iterations are repeated only twice in our experiments as described in
Section \ref{subsection:setting}.
}

\paragraph{Effect of Data Augmentation for CPS-SVM}

\begin{figure}[t]
  \begin{center}
    \includegraphics[width=\columnwidth]{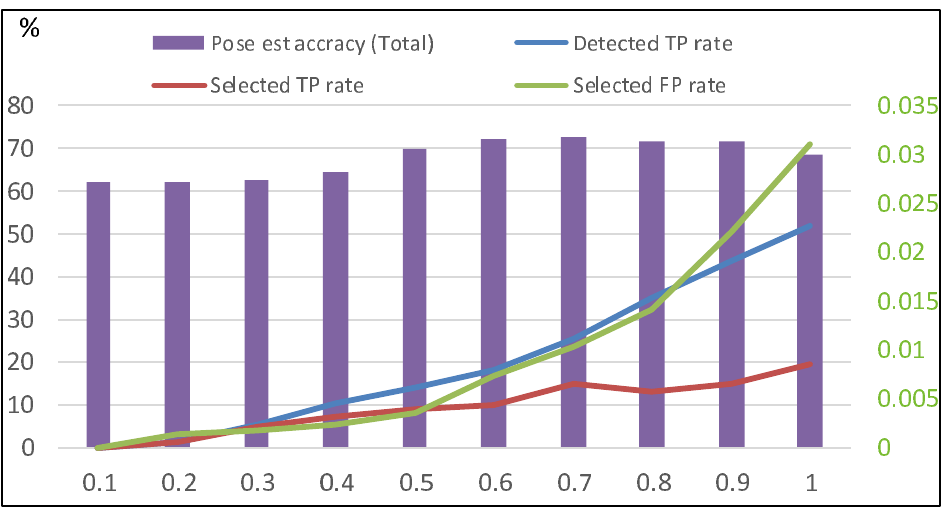}
    \caption{\textcolor{black}{
      Effects of parameter $\epsilon$, whose value is indicated
      in the horizontal axis. The lefthand vertical axis indicates the
      rates of detected and selected true-positives and the accuracy
      of pose estimation, while the righthand one indicates the rate
      of false-positives included in the selected poses.}}
    \label{fig:parameters}
  \end{center}
\end{figure}

The effect of parameter $\epsilon$ was examined with the test data of
the LSP (Figure \ref{fig:parameters}).
\textcolor{black}{
For simplicity, the results of only the complete learning scheme with
the training data of ``LSP'' (i.e., (i) Ours-weakC--1 in Table
\ref{table:results_lsp_pose}) are shown.
}
Compared with the growth of detected and selected true-positive poses
(indicated by blue and red lines, respectively, in the figure) with
increasing $\epsilon$, false-positives (indicated by a green line)
increases significantly. This may cause the decrease in the accuracy
of pose estimation (indicated by purple bars) at or above $\epsilon =
0.8$.

\begin{figure}[t]
  \begin{center}
    \includegraphics[width=.75\columnwidth]{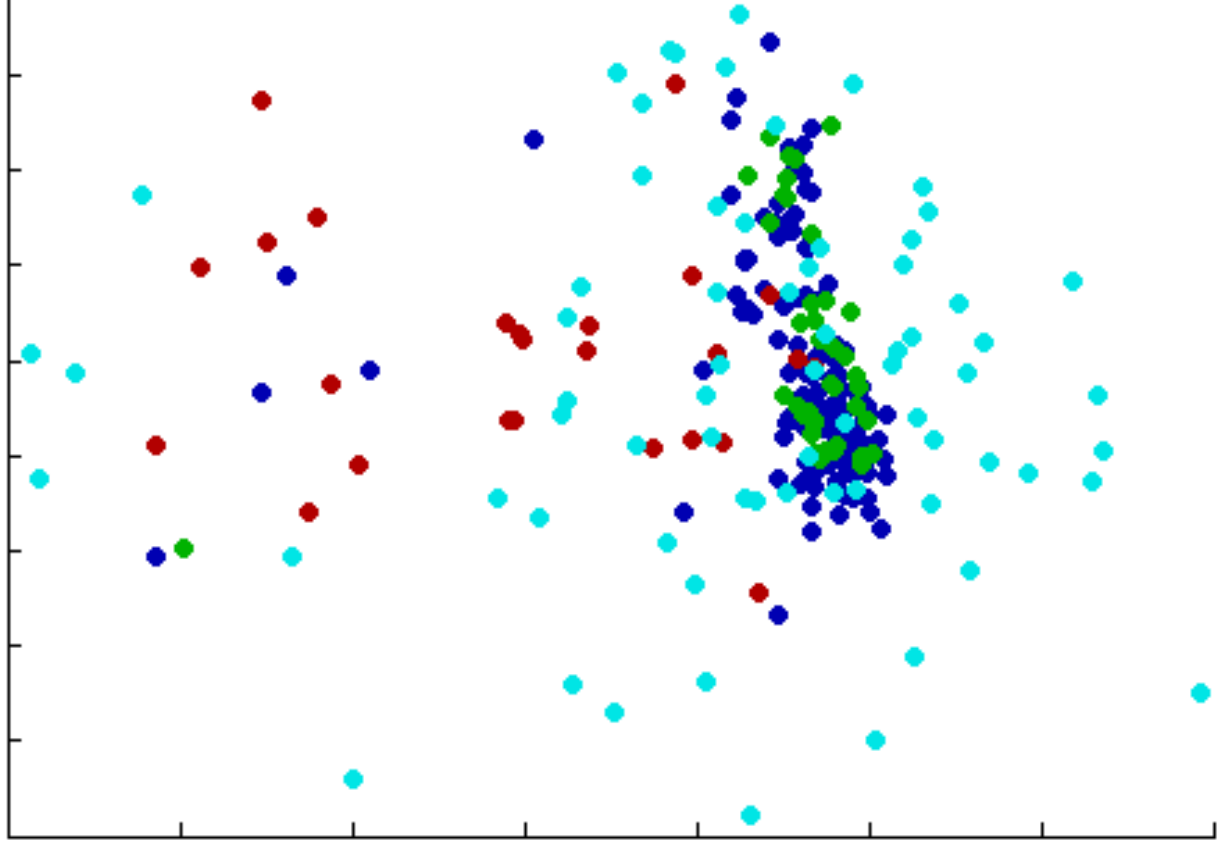}
    \caption{Distribution of PR features of
        training images (in ``soccer'' class of the LSP) . Blue,
        green, red, and skyblue points indicate annotated poses in
        the FS set, true-positives selected by the CPS-SVM,
        true-positives selected by clustering, and false-negatives,
        respectively. Note that distance between PR features in this
        2D space, given by PCA, is not identical to the one in the
        original PR feature space; even if two poses are
        closer/farther in this figure, they maybe farther/closer in the
        original PR feature space.}
    \label{fig:distribution}
  \end{center}
\end{figure}

\paragraph{Distributions of Detected True-positives}

For validating the effect of the semi- and weakly-supervised
learning scheme for selecting true-positives,
Figure \ref{fig:distribution} visualizes the
distribution of PR features in Ours-weakC--1.
While all true-positives selected by the CPS-SVM (indicated by green)
are close to poses in the FS set (indicated by blue), several
true-positives selected by clustering (indicated by red) are far from
the FS set as expected.

\paragraph{\textcolor{black}{More Unsupervised Data}}

\begin{table*}
  \begin{center}
    \caption{
      \textcolor{black}{
      Quantitative results of our semi-supervised training scheme
      using more unsupervised data obtained from the COCO 2016 keypoint
      challenge dataset. This scheme is evaluated with the test
      data of the LSP
      dataset and the strict PCK-0.2 metric \cite{DBLP:journals/pami/YangR13}.
      The results of (v) Baseline--2 \cite{wei2016cpm} and (w)
      Ours-semi--2 using LSP+LSPext+MPII are also shown for
      reference.
      }}
    \label{table:results_lsp_coco}
\begin{footnotesize}
    \begin{tabular}{l|cccccccc}
      \hline
      Method & Head & Shoulder & Elbow & Wrist & Hip & Knee & Ankle & Mean \\ \hline \hline
\multicolumn{9}{l}{\bf{LSP+LSPext+MPII}} \\ \hline
(v) Baseline--2~\cite{wei2016cpm}&97.8 &92.5 &87.0 &83.9 &91.5 &90.8 &89.9 &90.5 \\ \hline

(w) Ours-semi--2&92.4&80.8&70.3&65.7&82.5&73.3&68.8&76.3 \\ \hline \hline
\multicolumn{9}{l}{{\bf LSP+LSP+MPII+COCO}} \\ \hline
(z) Ours-semi--2&95.5&84.1&71.8&65.9&85.9&74.2&70.6&78.3 \\ \hline
    \end{tabular}
\end{footnotesize}
  \end{center}
\end{table*}

\textcolor{black}{
For our proposed scheme, unsupervised (US) data for
semi-supervised learning is much easier to
be collected than weakly-supervised (WS) data.
Here, the performance gain with more US set is investigated.
For the US set, the COCO 2016 keypoint challenge dataset
\cite{DBLP:conf/eccv/LinMBHPRDZ14} 
was used while no pose annotations in this dataset were used for our
experiments.
In total, over 126K human poses in the COCO were added to the US set.
}

\textcolor{black}{
The results are shown in Table \ref{table:results_lsp_coco}.
Compared with the huge number of the US set, the
performance gain is limited (i.e., $78.3 - 76.3 = 2.0$).
The performance might be almost saturated
because only human poses that are similar to those
included in the FS set can be detected and used for model re-learning
in our semi-supervised learning.
For more investigation, weakly-supervised learning using such a huge
training data is one of interesting future research directions while
we need action labels in the WS set.
}

\paragraph{\textcolor{black}{Qualitative Results}}

\begin{figure*}[t]
  \begin{center}
    \includegraphics[width=\textwidth]{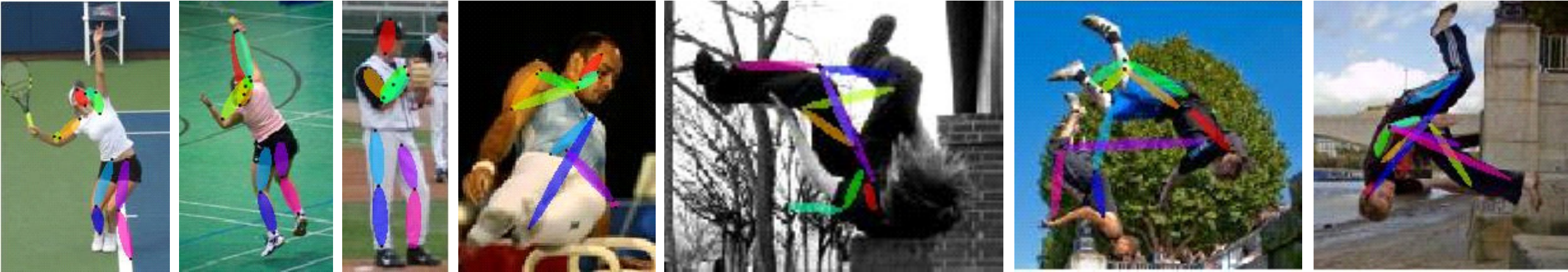}
    (j) Baseline--2 \cite{wei2016cpm} (HALF)

    \includegraphics[width=\textwidth]{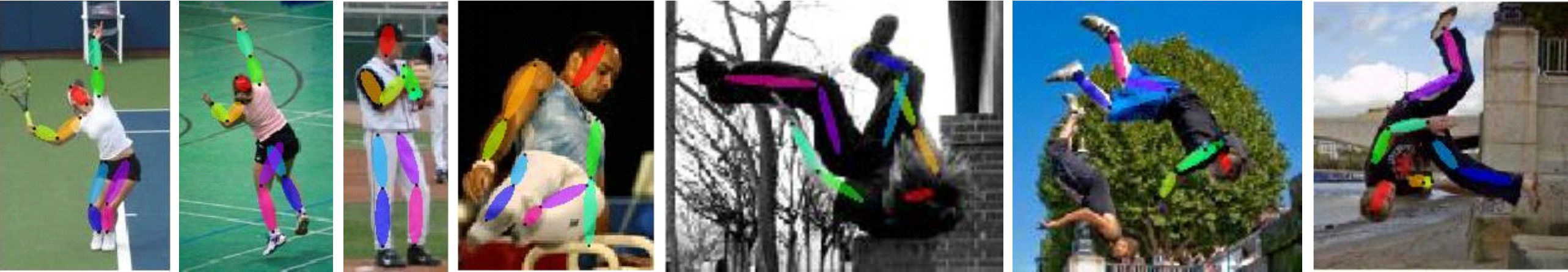}
    (y) Ours-weakC--2 (LSP+LSPext+MPII)

    \caption{\textcolor{black}{Improvement cases by our proposed method, Ours-weakC--2.}}
    \label{fig:visual_success}
  \end{center}
\end{figure*}

\begin{figure*}[t]
  \begin{center}
    \includegraphics[width=\textwidth]{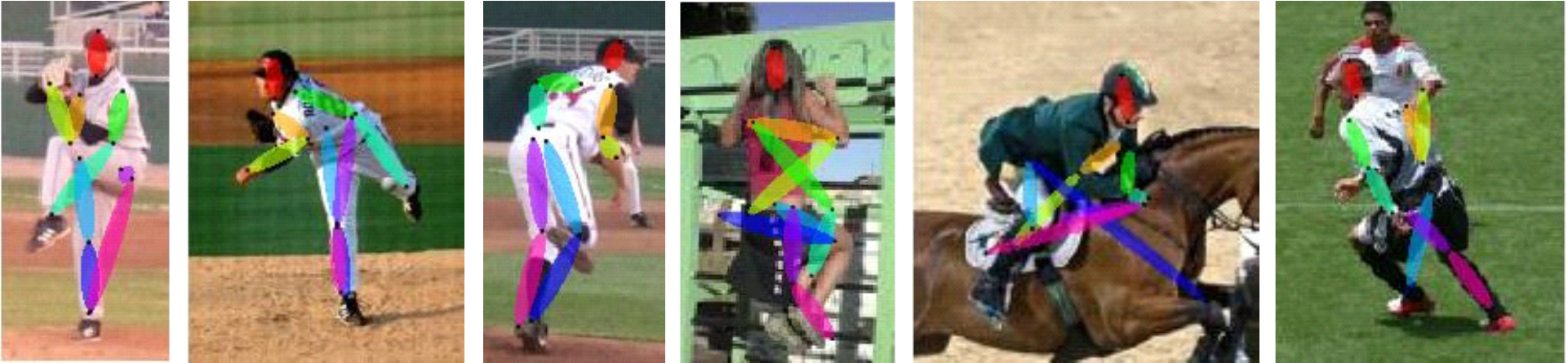}
    (1) \hspace*{14mm}(2) \hspace*{16mm}(3) \hspace*{15mm}(4)
    \hspace*{20mm}(5) \hspace*{20mm}(6)~~~~~

    (j) Baseline--2 \cite{wei2016cpm} (HALF)

    \includegraphics[width=\textwidth]{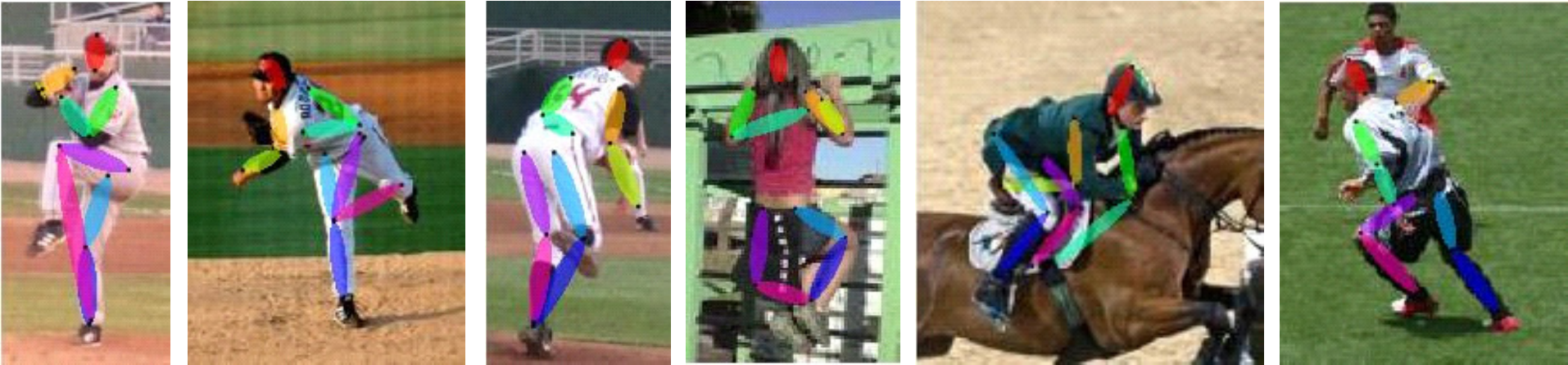}
    (1) \hspace*{14mm}(2) \hspace*{16mm}(3) \hspace*{15mm}(4)
    \hspace*{20mm}(5) \hspace*{20mm}(6)~~~~~

    (y) Ours-weakC--2 (LSP+LSPext+MPII)

    \caption{\textcolor{black}{Failure cases by our proposed method,
        Ours-weakC--2. The action class of each example is as
        follows. (1), (2), and (3): Baseball. (4): Parkour. (5):
        General. (6): Soccer.}}
    \label{fig:visual_failure}
  \end{center}
\end{figure*}

\textcolor{black}{
Several pose estimation results are shown in Figures
\ref{fig:visual_success} and \ref{fig:visual_failure}. In both
figures, Baseline--2 \cite{wei2016cpm} and our-weakC--2 are
trained by half
of LSP and by LSP+LSPext+MPII,
respectively; namely, the former and latter correspond to (j) and (y),
respectively.
In Figure \ref{fig:visual_success}, the results of all keypoints are
improved and localized successfully by our
method. In Figure \ref{fig:visual_failure}, on the other
hand, one or more keypoints are mislocalized by our method.
}

\textcolor{black}{
From the results in Figure \ref{fig:visual_failure}, we can find several
limitations of our proposed method.
In (1), (2), and (3), a pitching motion is observed. While a
large number of training data for this kind of motion is
included in ``baseball'' class,
body poses in this class are diversified (e.g., pitching, batting,
running, fielding).
That makes it difficult to model the pose variation in this class.
This difficulty can be possibly suppressed by more fine-grained
action grouping.
While (4) and (5) are ``parkour'' and ``general'',
respectively, these poses are not similar to any training samples in
their respective action class.
In (6), overlapping people make pose estimation difficult. For this
problem, a base algorithm should be designed for
multi-people pose estimation (e.g.,
\cite{DBLP:conf/eccv/InsafutdinovPAA16,cao2017realtime}).
}

\paragraph{More Quantitative Comparison}

\begin{table*}[t]
  \begin{center}
    \caption{Quantitative comparison using test data in the MPII
      dataset evaluated by PCKh-0.5
      \cite{DBLP:conf/cvpr/AndrilukaPGS14}.  Our proposed method
      (i.e., Ours-weakC--2) used
      9040 images in the MPII (i.e., half of the entire images)
      for the FS set and other images in
      ``LSP+LSPext+MPII'' dataset for the WS set. On the other hand,
      all images and
      annotations in MPII and ``LSP+LSPext+MPII'' were used for
      training in
      \cite{DBLP:conf/eccv/InsafutdinovPAA16,DBLP:conf/eccv/LifshitzFU16,DBLP:conf/eccv/GkioxariTJ16,bulat2016pose,newell2016eccv}
      (shown in the upper rows in the table) and
      \cite{pishchulin16cvpr,wei2016cpm} (shown in the lower rows),
      respectively. For reference, the results of the baseline
      \cite{wei2016cpm} that used only
      half of the entire images in the MPII (i.e., Baseline--2 (HALF)
      in the table)
      are shown.
      The best scores among supervised learning
      methods and
      methods that used only 9040 images for the FS set are colored by
      {\bf \textcolor{red}{red}} and {\bf \textcolor{blue}{blue}},
      respectively
}
    \label{table:results_mpii_pose}

\begin{footnotesize}
\begin{tabular}{ l|cccccccc }
\hline
Method & Head & Shoulder & Elbow & Wrist & Hip & Knee & Ankle & Mean \\ \hline \hline
\multicolumn{9}{l}{{\bf MPII}} \\ \hline
Insafutdinov et al. \cite{DBLP:conf/eccv/InsafutdinovPAA16} &97.4&92.7&87.5&84.4&{\bf \textcolor{red}{91.5}}&{\bf \textcolor{red}{89.9}}&{\bf \textcolor{red}{87.2}}&90.1 \\ \hline
Lifshitz et al. \cite{DBLP:conf/eccv/LifshitzFU16} &97.8&93.3&85.7&80.4&85.3&76.6&70.2&85.0 \\ \hline
Gkioxary et al. \cite{DBLP:conf/eccv/GkioxariTJ16}&96.2&93.1&86.7&82.1&85.2&81.4&74.1&86.1 \\ \hline
Bulat and Tzimiropoulos \cite{bulat2016pose}  &97.9&95.1&89.9&85.3&89.4&85.7&81.7&89.7 \\ \hline
Newell et al. \cite{newell2016eccv} &{\bf \textcolor{red}{98.2}}&{\bf \textcolor{red}{96.3}}&{\bf \textcolor{red}{91.2}}&{\bf \textcolor{red}{87.1}}&90.1&87.4&83.6&{\bf \textcolor{red}{90.9}} \\ \hline
Baseline--2~\cite{wei2016cpm} (HALF) &94.8&87.7&76.2&66.4&75.2&64.7&60.0&\underline{75.9} \\ \hline \hline
\multicolumn{9}{l}{{\bf LSP+LSPext+MPII}} \\ \hline
Pishchulin et al.~\cite{pishchulin16cvpr} &94.1 &90.2 &83.4 &77.3 &82.6 &75.7 &68.6 &82.4 \\\hline
Baseline--2~\cite{wei2016cpm} &97.8 &95.0 &88.7 &84.0 &88.4 &82.8 &79.4 &88.5 \\ \hline
Our-weakC--2 &\bf{\textcolor{blue}{96.9}}&\bf{\textcolor{blue}{92.9}}&\bf{\textcolor{blue}{84.6}}&\bf{\textcolor{blue}{78.6}}&\bf{\textcolor{blue}{84.6}}&\bf{\textcolor{blue}{75.8}}&\bf{\textcolor{blue}{70.9}}&\bf{\textcolor{blue}{\underline{83.4}}} \\ \hline
\end{tabular}
\end{footnotesize}
  \end{center}
\end{table*}

Pose estimation accuracy was evaluated also with the test data of the
MPII dataset (Table \ref{table:results_mpii_pose}).
For comparison, our semi- and weakly-supervised
learning scheme with clustering using
training images of ``LSP+LSPext+MPII'', which is equal to (y)
Ours-weakC--2 in Table \ref{table:results_lsp_pose},
is evaluated because it is the best among all of our proposed schemes.
Only half of training images in the MPII were used for the
FS set in our method.
While our method used a small amount of human pose
annotations (i.e., 13.7K, 29K, and 40K annotations in our method, MPII,
and ``LSP+LSPext+MPII'', respectively),
the effectiveness of semi- and weakly-supervised learning
is validated in comparison between our method and the base method
\cite{wei2016cpm} using only half of training data in the MPII (i.e., 83.4\%
vs. 75.9 \% in the mean, which are underlined in Table \ref{table:results_mpii_pose}).

\section{Concluding Remarks}
\label{section:conclusion}

We proposed semi- and weakly-supervised learning schemes for human pose
estimation.
While semi- and weakly-supervised learning schemes are widely used for
object localization and recognition tasks, this paper demonstrated
that such schemes are applicable to human pose estimation in still
images.
The proposed schemes extract correct poses from training images with
no human pose annotations based on (1) pose discrimination on the basis
of the configuration of all body parts, (2) action-specific models, and
(3) clustering and outlier detection using Dirichlet process mixtures.
These three functionalities allow the proposed semi- and
weakly-supervised learning scheme to outperform its baselines using
the same amount of human pose annotations.

Future work includes candidate pose synthesis and true-positive pose
selection using generative adversarial nets
\cite{DBLP:conf/nips/GoodfellowPMXWOCB14}, which can synthesize
realistic data from training data.
Since candidate pose synthesis and true-positive pose selection play
important roles in our proposed method, further improvement of these
schemes should be explored.

Experiments with more training data is also important. This
investigates and reveals the properties of the proposed schemes; for
example, (1) the relationship between the scale of the WS set and the
estimation performance and (2) the positive/negative effects of
true-positives/false-positives.
While our weakly-supervised learning scheme needs an action label in
each training image, unsupervised learning is more attractive for
increasing the amount of training images.
Automatic action labeling/recognition in training images
allows us to extend our weakly-supervised
learning to unsupervised learning.



\section*{References}

\bibliographystyle{elsarticle-num}

\end{document}